\documentclass[10pt,twocolumn,letterpaper]{article}
\usepackage{booktabs}       
\usepackage{amsfonts}       
\usepackage{nicefrac}       
\usepackage{microtype}      
\usepackage{bm}
\usepackage{amsmath}
\usepackage{algorithm}
\usepackage{pifont}
\usepackage{amssymb}
\usepackage{arydshln}
\usepackage{xspace}
\usepackage[utf8]{inputenc}
\usepackage{listings}
\usepackage{fontawesome}
\usepackage{tgpagella}
\usepackage{float}
\usepackage{multirow}
\usepackage{booktabs}
\usepackage[table]{xcolor}
\usepackage{siunitx}
\usepackage{caption}
\usepackage{hyperref}
\usepackage{makecell}
\usepackage{geometry}
\usepackage{caption}
\usepackage{subcaption}
\usepackage{adjustbox}
\usepackage{url}            
\usepackage{graphicx}
\usepackage{scalerel}
\usepackage{color}
\newcommand{\pub}[1]{{\color{gray}{\tiny{[{#1}]}}}}
\usepackage{hyperref}       
\usepackage{cvpr}              

\def\paperID{14275} 
\def\confName{CVPR}
\def\confYear{2024}

\title{DynFocus: Dynamic Cooperative Network Empowers LLMs with Video Understanding}
\author{Yudong Han$^{1\Diamond}$, Qingpei Guo$^{2\dag}$, Liyuan Pan$^{1 \dag}$, Liu Liu$^{3}$, Yu Guan$^4$, Ming Yang$^2$\\
\normalsize$^1$Beijing Institute of Technology, 
$^2$Ant Group,\\
\normalsize$^3$KooMap Dept., Huawei,
\normalsize$^4$University of Warwick \\
{\tt\small hanyudong.sdu@gmail.com, qingpei.gqp@antgroup.com, liyuan.pan@bit.edu.cn} \\
{\tt\small liuliu33@huawei.com, Yu.Guan@warwick.ac.uk, m.yang@antgroup.com}
}
\begin{document}

\maketitle

\begin{abstract}
The challenge in LLM-based video understanding lies in preserving visual and semantic information in long videos while maintaining a memory-affordable token count. However, redundancy and correspondence in videos have hindered the performance potential of existing methods. Through statistical learning on current datasets, we observe that redundancy occurs in both repeated and answer-irrelevant frames, and the corresponding frames vary with different questions. This suggests the possibility of adopting dynamic encoding to balance detailed video information preservation with token budget reduction.
To this end, we propose a dynamic cooperative network, DynFocus, for memory-efficient video encoding in this paper. Specifically, i) a Dynamic Event Prototype Estimation (DPE) module to dynamically select meaningful frames for question answering; (ii) a Compact Cooperative Encoding (CCE) module that encodes meaningful frames with detailed visual appearance and the remaining frames with sketchy perception separately. We evaluate our method on five publicly available benchmarks, and experimental results consistently demonstrate that our method achieves competitive performance. Code is available at \color{magenta}{\url{https://github.com/Simon98-AI/DynFocus}}
\end{abstract}
\begin{figure}[ht]
\centering
 \subfloat[Redundancy]{
\hspace{-1.5em} 
\includegraphics[scale=0.45]{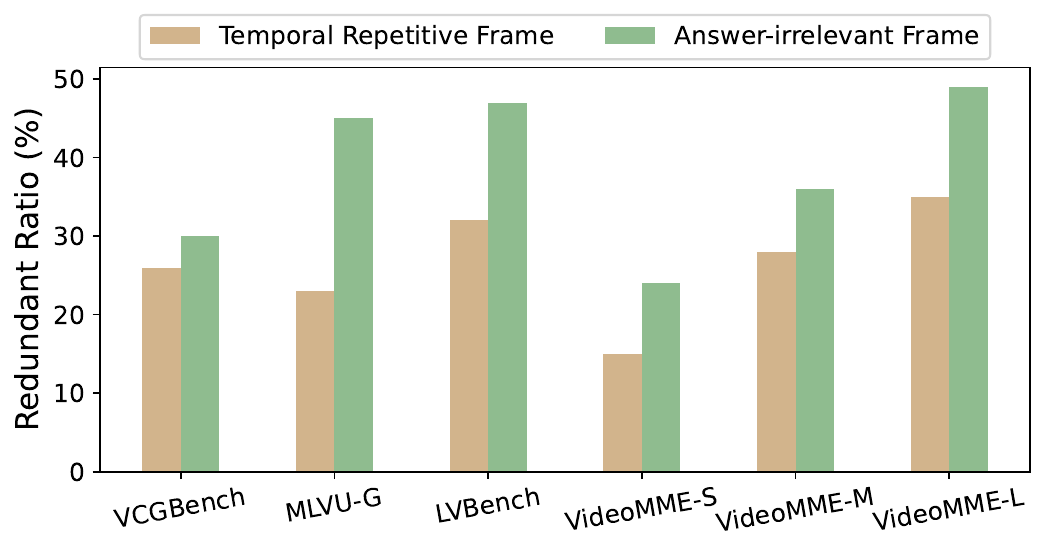}}
\vspace{-1.5em}
\subfloat[Correspondence]{
\includegraphics[scale=0.32]{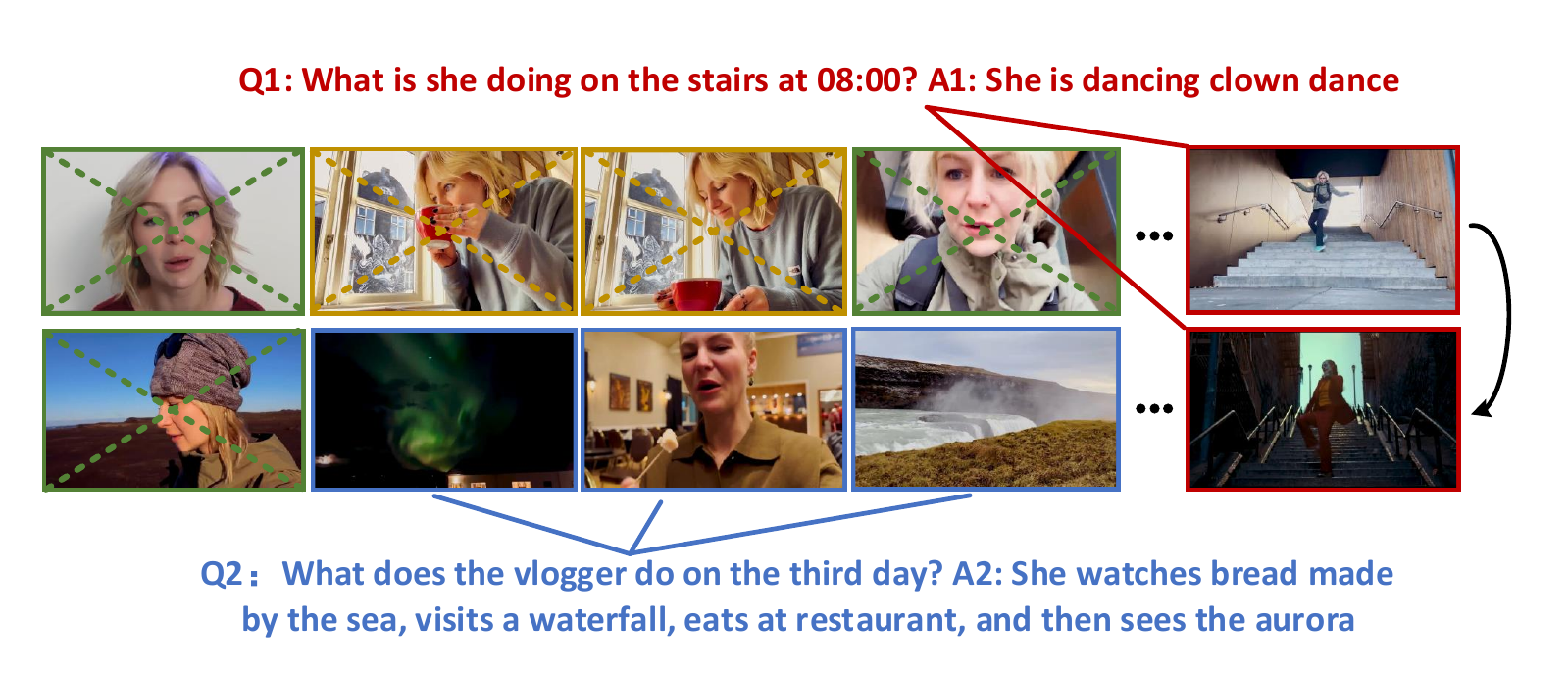}}\vspace{-0.1cm} 
\caption{
Concept of redundancy and correspondence in our pipeline. (a) The proportion of redundancy for video datasets\protect\footnotemark. Redundancy includes both repeated and answer-irrelevant frames. Repeatance gauges the redundancy between consecutive frames, while answer-irrelevance refers to frames with a marginal contribution to question answering. 
(b) An example of correspondence. Given a video, we highlight the corresponding question/answer pairs and frames using red and blue boxes, respectively. 
}
\vspace{-1.5em}
\label{fig:intro}
\end{figure}
\footnotetext{The implementation is provided in supplementary materials.} 

\renewcommand{\thefootnote}{ } 
\footnotetext{$\dag$ Corresponding authors.~~$\Diamond$ Work done during internship at Ant Group.}
\section{Introduction}
Large Language Models (LLMs) have shown their ability on general AI
~\cite{DBLP:journals/corr/abs-2001-08361}. Vision Language Models (VLMs) extend the capabilities of LLMs to process visual data, demonstrating proficiency in tasks such as image captioning and visual question answering. However, challenges arise in video understanding, especially with long-term videos, where representing consecutive video frames requires an excessive number of tokens, leading to high memory usage. 

Recent attempts use average pooling, attention, or dynamic masking to reduce video tokens spatially~\cite{li2023llama, maaz2023video, DBLP:journals/corr/abs-2405-07798, jin2023chat, DBLP:journals/corr/abs-2404-00308}. However, redundant frames lead to the neglect of key visual details. Several works~\cite{DBLP:conf/cvpr/0004LJJCSSL24, videollamb} capture visual appearance with memory banks to preserve key details. However, the key details vary in correspondence to questions, which can easily result in the loss of keyframes from long videos and increase the overhead of the memory banks.

In Fig.~\ref{fig:intro}, we illustrate examples of redundancy and correspondence in videos. We observe that i) there is significant redundancy among frames, with only a few meaningful frames directly contributing to question answering. This suggests the potential for adopting a dynamic frame encoding strategy to reduce tokens based on their contribution. ii) Answering different questions generally requires focusing on different parts of the frame. Therefore, dynamically identifying meaningful frames offers better flexibility for sophisticated video content understanding.

In this paper, we propose a dynamic cooperative framework, \textit{DynFocus}, for memory-efficient video encoding. Specifically, it consists of two key modules: {Dynamic Event Prototype Estimation} (DPE) and {Compact Cooperative Encoding} (CCE). DPE serves as the dynamic selector to accurately discern the meaningful frames, which takes both redundancy and correspondence to question answering into consideration as the selection standard. 
Afterwards, CCE complies with the dynamic encoding principle. The meaningful frames are encoded with fine-grained context features for detailed visual appearance, whereas those redundant frames are substantially encapsulated into a few tokens for sketchy perception, which enables LLM to capture broader temporal clues within a fixed receptive field. These two modules reconcile the nuanced visual-semantic understanding with affordable token quota.


Moreover, our CCE module draws inspiration from the cooperation of retinal ganglion cells in the primate visual system. Biological studies~\cite{nature, visual-iovs} found that in these cells, \textit{Rod} cells perceive the overall scene in a wide field of view, while \textit{Cone} cells understand complex scenes with fine details. These cells are located at the periphery of the retina and arranged in a parallel manner, receiving the signals but activated under different conditions. Our framework is analogous in two aspects: (1) Which cell is activated depends on whether the current input frame is meaningful or not. (2) The meaningful frames are encoded with fine-grained tokens as key detailed clues, akin to $\textit{Cones}$, whereas the marginal frames are condensed into low-resolution tokens, ensuring better temporal consistency, similar to $\textit{Rods}$. We hope these relations will further support the design philosophy of our method and reveal its rationality.

Our contributions are summarized as follows:
\begin{itemize}
\item We propose a dynamic cooperative network, $\textit{DynFocus}$, towards memory-efficient video encoding within LLM, inspired by the biological concept of $\textit{Cone}$ and $\textit{Rod}$ cells. 
\item We introduce two modules, DPE and CCE, that dynamically balance subtle visual appearance with sketchy temporal perception using affordable tokens.
 \item Experimentally, we achieve the competitive even SOTA performance on two publicly mainstream short video benchmarks, three long video benchmarks, and one diagnosis benchmark on video hallucination. 
\end{itemize}

\section{Related Work}

\textbf{Video-based Large Language Models.} In recent years, Vision Language Models (VLMs) has emerged to extend the capabilities of LLMs~\cite{radford2018improving, chowdhery2023palm, touvron2023llama, vicuna2023, openai2022chatgpt} to handle diverse and complicated inputs with satisfactory generalization. Generally, VLMs incorporate additional connector to bridge the semantic gap between input video content and LLMs, further performing modality alignment and instruction tuning on video-based dataset. However, video understanding presents the significant challenges due to their extensive memory overhead. Several studies have dedicated to addressing these challenges with greater efficiency. Video-ChatGPT~\cite{maaz2023video} adopts both spatial and temporal pooling to condense video tokens. VideoChat~\cite{li2023videochat} employs a learnable Q-former~\cite{DBLP:conf/nips/Dai0LTZW0FH23} to aggregate the similar tokens for memory reduction. Chat-UniVi~\cite{jin2023chat} develops a unfied framework for processing both image and video, which reduces spatial and temporal tokens through multi-stage token merging. Although these methods alleviate the memory usage to some extent, they often discard the abundant temporal clues by sampling parts of frames as the input. To compensate the loss of temporal clues, LLaMA-VID~\cite{li2023llama} innovates with a dual-token approach that represents each frame with context and content tokens, which allows for larger video throughput. MovieChat~\cite{Song2023MovieChatFD} incorporates the short-term memory and long-term memory into unified framework, strategically combining similar frames to reduce memory footprint while capture the temporal clues. Similarly, MA-LLM~\cite{he2024malmm} stores past video information in a memory bank, which allows to reference historical video content for long-term analysis without exceeding memory limits. However, these methods exhibit proficiency in capturing temporal clues at the expense of discarding the visual details. In a nutshell, they struggle to jointly capture the spatial details and temporal dynamics effectively.

\noindent{\textbf{Dynamic Networks.}} Dynamic networks, adjusting the encoding strategy according to specific input, have recently garnered the burgeoning interest across various domains. Early methods mainly focus on traditional image classification by channel pruning or layer skipping. For example, BlockDrop~\cite{DBLP:conf/cvpr/WuNKRDGF18} designed an auxiliary policy network to determine whether skip or execute convolutional blocks via reinforcement optimization. Based on dynamic mechanism, a series of research efforts are devoted to better adapting to the various dynamic scenes. Specifically, Dynamic~\cite{DBLP:conf/cvpr/LiSCLZWS20} proposes a routing network with soft conditional gate to adaptively search data-dependent scale transformation paths for semantic segmentation. In the field of image question answering, SUPER~\cite{DBLP:journals/tip/HanYWWN23} develops a semantic-aware modular routing framework to recursively handle different complexity of visual scene. In this work, we marks the first attempt to reveal the substantial potential of dynamic encoding strategies when understanding the complicated long-term video.

\begin{figure*}
    \centering
\includegraphics[width=.96\linewidth]{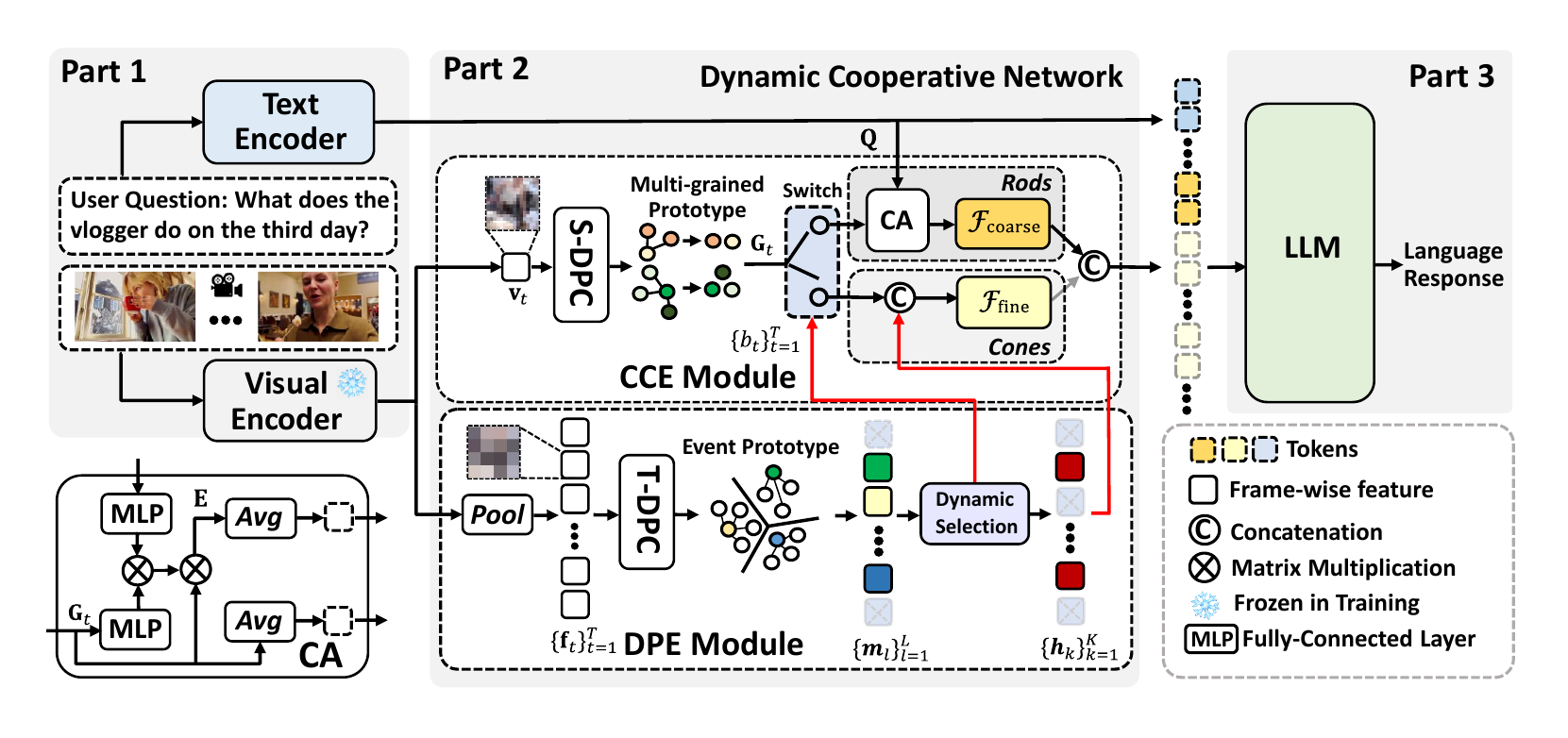}
    \vspace{-2em}
    \caption{Schematic Illustration of \textit{DynFocus}. Our method takes the user instruction and video frames as input, and yields the compact video tokens from CCE module for LLM. Specifically, DPE module serves as the selector to identify the prototypes that contribute greatly to answer, providing CCE module with event prototype $\{\mathbf{h}_{k}\}_{k=1}^{K}$ and the binary mask $\{b_{t}\}_{t=1}^{T}$, which is marked with two red arrows. Benefited from this, CCE module dynamically encode the critical prototypes with more tokens, and encapsulate the marginal prototypes with few tokens. T-DPC and S-DPC represent the DPC-KNN clustering temporally and spatially, respectively.} 
    \label{fig:framework}
    \vspace{-1em}
\end{figure*}

\section{Dynamic Cooperative Network}
As shown in Fig~\ref{fig:framework}, the overall framework is comprised of three parts. \textbf{Part 1:} visual and text encoder are adopted to produce the corresponding features; \textbf{Part 2:}  the proposed dynamic cooperative network serves as the connector to compress the video content for LLM, which consists of two modules, DPE module and CCE module; \textbf{Part 3:}  the foundational LLM receives the token sequence outputted from {Part 2} to generate the language response.

\subsection{Visual and Text Encoder}
Given a $T$-frame video, we extract the frame-wise features 
$\mathbf{V}=\{\mathbf{v}_{t}\}_{t=1}^{T}$,
using a pre-trained visual encoder, where $\mathbf{v}_{t} \in \mathbb{R}^{N \times d}$ denotes the feature of the $t$-th frame. Here, $N$ is the number of image patches and $d$ is the feature dimension.
For the text encoder, user instruction is fed to the pre-trained
text encoder to generate the text features $\mathbf{Q}=\{\mathbf{q}_{r}\}_{r=1}^{R}$,
where $\mathbf{q}_{r} \in \mathbb{R}^{d^{\prime}}$ denotes the feature embedding of each token in user instruction. $d^{\prime}$ is the feature dimension and $R$ is the total number of tokens. 

\subsection{Dynamic Event Prototype Estimation}~\label{sec::epe}
Given frame-wise features $\mathbf{V}$, we aim to estimate event prototypes, \ie, discriminative features that are most relevant to ground-truth answer.

For each frame feature $\mathbf{v}_{t}$, we first perform local average pooling spatially. The pooling process reduces the number of features from $N$ to $P$ for efficiency, resulting in  $\mathbf{f}_{t} = \mathrm{Pool}(\mathbf{v}_{t}), \mathbf{f}_{t} \in \mathbb{R}^{P \times d}$. Redundancy exists in $T$-frame features $\mathbf{F}=\{\mathbf{f}_{t}\}_{t=1}^{T}$. To remove redundancy and identify representative features from $\mathbf{F}$, we perform clustering on $\mathbf{F}$, and use cluster centers to estimate event prototypes.

Following~\cite{jin2023chat}, we borrow components from the traditional DPC-KNN~\cite{DBLP:journals/kbs/DuDJ16} algorithm for obtaining cluster centers. 
For self-contain purposes, we briefly summarize two important variables (local density $\rho_{t}$ and distance indicator $\delta_{t}$ ) in DPC-KNN below.

For clustering, the local density $\rho_{t}$ measures the mean distance to $C$ nearest neighbors of the $t$-th frame, and is given by,
\begin{align}
\rho_{t} = \mathrm{exp}\left(-\frac{1}{C} \sum_{{t^{\prime}} \in \mathcal{N}({t})} \frac{1}{P} \Vert \mathbf{f}_{t} - \mathbf{f}_{t^{\prime}} \Vert_{F}^{2}\right),
\label{eq:1}
\end{align}
where ${{t^{\prime}} \in \mathcal{N}({t})}$ denotes that $\mathbf{f}_{t^{\prime}}$ is in the neighborhood of $\mathbf{f}_{t}$.
$\left\| \cdot \right\|_{F}$ denotes the Frobenius norm, and it is also used to perform $C$ nearest neighbor search for each $\mathbf{f}_{t}$.

The distance indicator $\delta_{t}$ measures the possibility of $t$-th frame to be cluster center by calculating the minimum distance between $\mathbf{f}_{t}$ and any other frames with higher density, and is given by, 
\begin{align}
\delta_{t}=\left\{\begin{array}{l}
 \underset{t^{\prime}}{\min} \frac{1}{P} \left\|\mathbf{f}_{t}-\mathbf{f}_{t^{\prime}}\right\|^{2}_{F}, \text { if } \exists t^{\prime} \text { s.t. } \rho_{t^{\prime}} \textgreater \rho_{t} \\
 \underset{t^{\prime}}{\max} \frac{1}{P} \left\|\mathbf{f}_{t}-\mathbf{f}_{t^{\prime}}\right\|^{2}_{F}, \text { otherwise }.
\end{array}\right.
\label{eq:2}
\end{align}

We use the product of $\rho_{t}$ and $\delta_{t}$ to measure the importance of each frame. Frames with high scores are more likely to be informative. We first sort scores $\{\rho_{t} \times \delta_{t}\}_{t=1}^{T}$ in the decreasing order and then take the Top-$L$ frame features from $\mathbf{F}$ with high scores as representative features. Finally, we normalize representative features using their importance scores and estimate event prototypes as follows,
\begin{equation}
\mathbf{m}_{l} = \frac{\sum_{t^{\prime} \in \mathcal{N}(l)} \mathrm{exp}(\rho_{t^{\prime}} \cdot \delta_{t^{\prime}}) \mathbf{f}_{t^{\prime}}}{\sum_{t^{\prime} \in \mathcal{N}(l)}  \mathrm{exp}(\rho_{t^{\prime}} \cdot \delta_{t^{\prime}})}, l\in[1,L],
\label{eq:3}
\end{equation}
where $\mathrm{exp}(\rho_{t^{\prime}} \cdot \delta_{t^{\prime}})$ denotes the importance weight of $\mathbf{f}_{t^{\prime}}$.

Note that the obtained event prototypes $\mathbf{M}=\{\mathbf{m}_{l}\}_{l=1}^{L}$ from Eq.~\eqref{eq:3} are estimated only based on frame features, and are not aligned to the ground-truth answer. In other words, $\mathbf{m}_{l} \in \mathbb{R}^{P \times d}$ may contain visual redundancy, which would disturb useful clues for sophisticated video understanding. In light of this, we aim to further select answer-relevant event prototypes from $\mathbf{M}$. Specifically, we use a multi-layer perceptron (MLP) network $\mathcal{U}(\cdot)$ to regress frame-wise scores. Note that $\mathcal{U}(\cdot)$ is learned end-to-end with the supervision from LLM, thus is aligned to ground-truth answer implicitly. The regression is given by,
\begin{align}
s_{l} = \mathcal{U}\left( \mathrm{Max}(\mathbf{m}_{l}) || \mathrm{Avg}(\mathbf{m}_{l}) \right),
\end{align}
where $\mathrm{Max}(\mathbf{m}_{l})  \in \mathbb{R}^{d}$ denotes the row feature with maximum feature norm ($L_2$) across $P$ rows. $\mathrm{Avg}(\mathbf{m}_{l})\in \mathbb{R}^{d}$ denotes the averaged row feature across $P$ rows. $\cdot||\cdot$ denotes concatenation. After collecting all scores into a score vector $\mathbf{s}=\{\mathbf{s}_{l}\}_{l=1}^{L}$, we perform min-max normalization to normalize score values to be within $[0,1]$. Finally, we sort scores in the decreasing order and take the Top-$K$ event prototypes from $\mathbf{M}$ with high scores, indicated by the index vector $\mathbf{p}$, as filtered event prototypes. The filtered event prototypes and the index vector are denoted as $\mathbf{H} = \{\mathbf{h}_{k}\}_{k=1}^{K}$ and $\mathbf{p} = \mathrm{topk}(\mathbf{s}) \in \mathbb{N}^{K}$, respectively.

We also retrieve indices of filtered event prototypes in the $T$-frame video, obtaining the binary index mask $\mathbf{b} = \{b_{t}\}_{t=1}^{T}$. Notably, $b_{t}=1$ indicates important frame, while $b_{t}=0$ signifies redundant frame.

\vspace{+0.5mm}
\noindent{\bf Training $\mathcal{U}(\cdot)$.} Note that the Top-$K$ operation is not differentiable and thus stops the gradient propagation from LLM to update our score net $\mathcal{U}(\cdot)$. This limitation restricts our $\mathcal{U}(\cdot)$ to dynamically estimate the redundancy and flexibly capture correspondence without auxiliary loss supervision. To address this issue, we transform the Top-$K$ operation into solving a linear programming problem to make our network end-to-end trainable. Specifically, we convert the index vector $\mathbf{p} = \left[ p_{1}, ..., p_{K}\right]$ into a stack of $L$ one-hot vector with $K$ elements, denoted as $\mathbf{P}= \left[ \mathbf{I}_{p_{1}}, ..., \mathbf{I}_{p_{K}}\right] \in \{0, 1\}^{L \times K}$. Here, $\mathbf{I}_{p_{1}}$ denotes the one-hot vector where only the $p_{1}$-th element is set to 1. As a result, the filtered event prototypes with top-$K$ scores can be summarized as $\mathbf{H} = \mathbf{P}^\top\mathbf{M}$. Afterwards, we resort to the perturbed maximum method~\cite{DBLP:journals/corr/abs-2002-08676} to construct a differentiable operator. In theory, selecting top-$K$ prototypes via subspace projection matrix $\mathbf{P}$ equals to solving a linear programming problem , 
\begin{align}
     \mathrm{argmax}_{\mathbf{P} \in \mathcal{C}} \left \langle \mathbf{P}, \mathbf{s}\mathbf{1}^\top \right \rangle,
     \label{eqn:9}
\end{align}
where $\mathbf{s} \mathbf{1}^\top \in \mathbb{R}^{K \times L}$ denotes the score vector $\mathbf{s}$ replicated $L$ times, $ \left \langle  \right \rangle$ denotes the flatten operation followed by dot product. $\mathcal{C}$ is the orthodox convex polytope constrain set $\mathcal{C} = \{ \mathbf{P} \in \mathbb{R}^{K\times L}: \mathbf{P}_{k,l} \geq 0, \mathbf{1}^\top \mathbf{P} = \mathbf{1}\}$. We follow ~\cite{DBLP:journals/corr/abs-2002-08676} to perform forward and backward operations to solve $\mathbf{P}$.
Specifically, solving Eqn.~\ref{eqn:9} could be achieved by taking the expectation of random perturbations, 
\begin{align}
     \mathbf{P}_{\sigma} = \mathbb{E}_{P} \left[ \mathrm{argmax}_{\mathbf{P} \in \mathcal{C}} \left \langle \mathbf{P}, \mathbf{s}\mathbf{1}^\top + \sigma \mathbf{Z} \right \rangle \right],
\label{eqn:6}
\end{align}
where $\mathbf{Z}$ is a perturbed vector sampled from the uniform Gaussian distribution and $\sigma$ serves as the hyper-parameter to control the variance of perturbation. Following~\cite{8093856}, the Jacobian associated with Eqn.~\ref{eqn:6} can be simplified as,
\begin{align}
     \frac{\partial \mathbf{P}_{\sigma}}{\partial \mathbf{s}} = \mathbb{E}_{P} \left[ \mathrm{argmax}_{\mathbf{P} \in \mathcal{C}} \left \langle \mathbf{P}, \mathbf{s}\mathbf{1}^\top + \sigma \mathbf{Z} \right \rangle  \mathbf{Z}
 / \sigma \right],
 \label{eqn:7}
\end{align}
By means of Eqn.~\ref{eqn:7}, the gradient from autoregressive loss in LLM would update the distribution of representation $\mathbf{H}$ and matrix $\mathbf{P}$, thereby updating our score network $\mathcal{U}(\cdot)$ via $\frac{\partial \mathbf{P}_{\sigma}}{\partial \mathbf{s}}$ according to the chain rule. As a result, our DPE module can be trained together with LLM in an end-to-end fashion, which effectively mitigates the answer-irrelevant visual nuisance in video while achieving dynamic selection in accordance with answer and question.

\subsection{Compact Cooperative Encoding}
Given frame-wise features 
$\mathbf{V}=\{\mathbf{v}_{t}\}_{t=1}^{T}$, event prototypes $\mathbf{H} = \{\mathbf{h}_{k}\}_{k=1}^{K}$, and their corresponding index mask $\mathbf{b} = \{b_{t}\}_{t=1}^{T}$, we perform cooperative encoding for memory-efficient video understanding. Frames that contribute greatly to answer (\ie, $b_t=1$) will be encoded with more tokens than marginal frames (\ie, $b_t=0$), to capture intricate spatial details. 

For each frame feature $\mathbf{v}_{t}$, we use the same clustering pipeline (Eq.~\eqref{eq:1},\eqref{eq:2},\eqref{eq:3}) in Sec.~\ref{sec::epe} to estimate spatial object prototypes $\mathbf{Z}_{t}=\{\mathbf{z}_{t,i}\}_{i=1}^{I}$. The only difference is that spatial clustering is performed on $N$ patch features within single frame feature $\mathbf{v}_{t}$, aggregating $N$ patch features into $I$ prototypes. In contrast, temporal clustering in Sec.~\ref{sec::epe} is performed on $T$ frame features in the whole video.

We find that semantic abstraction of $\mathbf{v}_{t}$ can be achieved when spatial clustering is performed multiple times. For example, concepts such as ``person" and ``dog" are progressively formed from the low-level attribute or color information. We thus use multiple clustering layers to capture more abundant visual details. The output spatial object prototypes of layer $j$ is fed to layer $j+1$ for abstraction, which means that number of prototypes participating in subsequent layers reduce progressively.

Collecting all layer outputs results in our multi-grained spatial object prototypes $\mathbf{G}_{t} = \{\mathbf{Z}^{(j)}_{t}\}_{j=1}^{J}$. Here, $J$ is the total number of clustering layers.

\vspace{-4mm}
\paragraph{\textit{Cones} Encoding.} We mimic cones to focus on fine visual appearance. Specifically, frames with $b_{t}=1$ are encoded with the combination of event and its corresponding multi-grained spatial prototypes,
\begin{equation}
\mathbf{U}_{t,b_{t}=1} = \mathcal{F}_\text{fine}\left ( \mathbf{h}_{t} || \mathbf{G}_{t}  \right ),
\end{equation}
where $\mathcal{F}_\text{fine}$ is a simple MLP network. Note that no feature pooling operation is performed to capture delicate details, and the number of tokens in $\mathbf{U}_{t,b_{t}=1}$ equals the number of summation of event and multi-grained spatial prototypes.

\vspace{-4mm}
\paragraph{\textit{Rods} Encoding.} We mimic rods to focus on coarse temporal dynamics towards broader video understanding. Specifically, frames with $b_{t}=0$ are encoded with the modulation of text embedding $\mathbf{Q}$, to obtain text-grounded visual clues,
\begin{equation}
\mathbf{E} = \mathrm{Softmax} \Bigg ( \frac{ f_{q}(\mathbf{G}_{t}) (f_{k}(\mathbf{Q}))^{\top}}{\sqrt{d}} \Bigg ) \mathbf{G}_{t},
\end{equation}
where $f_{q}(\cdot)$ and $f_{k}(\cdot)$ represent the linear projection, which map the spatial object prototypes and textual embedding into query and key, respectively.

To facilitate the memory-efficient video understanding, we condense $\mathbf{E}$ to a single token using average pooling. Combining global content token, we extract compact embedding, 
\begin{equation}
\mathbf{U}_{t, b_{t}=0} =  \mathcal{F}_\text{coarse}\left ( \mathrm{Avg}(\mathbf{E}) || \mathrm{Avg}(\mathbf{G}_{t})  \right ),
\end{equation}
where $\mathcal{F}_\text{coarse}$ is a simple MLP network, and $\mathbf{U}_{t, b_{t}=0} \in \mathbb{R}^{2d^{\prime}}$. Due to that $\mathbf{U}_{t, b_{t}=0}$ only has two tokens, it enables smooth temporal transition and improves the scene consistency for consecutive frames.  
\vspace{-4mm}
\paragraph{Cooperative Encoding.} Given embeddings $\mathbf{U}_{t,b_{t}=1}$ and $\mathbf{U}_{t,b_{t}=0}$ from $\textit{Cones}$ and $\textit{Rods}$, respectively, we combine them in a token-wise manner to obtain the dynamic embedding of the $t$-th frame,
\begin{equation}
\mathbf{O}_{t} = b_{t} \cdot (\mathbf{U}_{t,b_{t}=1} || \mathbf{U}_{t,b_{t}=0}) + (1-b_{t}) \cdot \mathbf{U}_{t,b_{t}=0}.
\end{equation}

The video embedding $\mathbf{O} = \{\mathbf{O}_{t}\}_{t=1}^{T}$ and the text embedding $\mathbf{Q}$ are translated into the language space in token format, which is used to generate response from LLMs.

\subsection{Training Strategy}
In this work, we adopt a two-stage training scheme following previous work~\cite{li2023llama}.

\noindent\textbf{Stage1: Vision-Language Alignment.}
In the first stage, we pre-train our dynamic cooperative network while freezing both the visual encoder and LLM. It is noteworthy that we only preserve the parameter of projector $\mathcal{F}_{\text{fine}}(\cdot)$ and $\mathcal{F}_{\text{coarse}}(\cdot)$ as the initialization in the second stage. Freezing LLM in the first stage is crucial to effectively align the representation space between video content and language without sacrificing any discernible performance of LLMs.

\noindent\textbf{Stage2: Instruction Tuning.}
After the first stage, the model possesses the ability of understanding the image within the language space, but fails to flexibly generate the reasonable and coherent linguistic responses. Therefore, in the second stage, we fully fine-tune the LLM and overall parameters in DPE module and CCE module on a instruction-following dataset. This dataset is a composite of pure text QA pairs, single- or multi-turn image QA pairs, and video QA pairs presented in a conversational format. In terms of instruction formulation, different formats are adopted for different kinds of input, and input $\mathrm{\left \langle prompt \right \rangle}$ vary with datasets. Meanwhile, the image token $\mathrm{\left \langle image \right \rangle}$ denotes the placeholder of image or videos, which is randomly inserted at the beginning or end of user prompt or question when training.

\section{Experiments}
\subsection{Experimental Setup}
\noindent\textbf{Implementation Details.}
We use the pre-trained ViT-G/14 from EVA-CLIP~\cite{DBLP:conf/cvpr/FangWXSWW0WC23} as the visual encoder to extract the features of each frame in video, and it can be further changed to other clip-based video encoders. We use pre-trained Qformer weight from InstructBLIP~\cite{DBLP:conf/nips/Dai0LTZW0FH23} as the textual encoder. Besides, we adopt the Vicuna-7B-1.5 model~\cite{vicuna2023} as our foundational LLM. Our model is trained using 8 $\times$ NVIDIA A100 80G GPUs. See more details in the supplementary material.

\noindent\textbf{Training Datasets.} We leverage image-video joint training following most of works to enhance the multi-modality understanding of LLMs. Specifically, we leverage the image-to-text dataset LLaVA-filter-CC3M~\cite{DBLP:conf/acl/SoricutDSG18} image-caption pairs for the first stage training following LLaVA-VID~\cite{li2023llama}, and LLaVA-665K~\cite{DBLP:journals/ijcv/GoyalKASBP19, hudson2019gqa, DBLP:conf/emnlp/KazemzadehOMB14, krishna2017visual_genome, liu2023improvedllava, Mao2015GenerationAC, mishra2019ocr, schwenk2022okvqa, sidorov2020textcaps} image QA pairs and ScienceQA~\cite{lu2022learn} for the second stage training, respectively. For video-to-text dataset preparation, we use WebVid-2.5M~\cite{bain2021frozen} video-caption pairs for the first stage, and a subset from VideoChat2 for the second stage, including VideoChatGPT-100K~\cite{maaz2023video}, WebVid-10M-QA~\cite{bain2021frozen}, NExT-QA~\cite{DBLP:conf/iccv/BainNVZ21}, and CLEVRER~\cite{DBLP:conf/iclr/YiGLK0TT20}. And all the samples are formulated as the uniform input format as LLaMA-VID~\cite{li2023llama}. 

\begin{table}[t]
\setlength{\tabcolsep}{3pt}
\centering
    \caption{Performance comparisons on zero-shot QA benchmark, including MSVD-QA \cite{wu2017deep}, MSRVTT-QA \cite{xu2016msr}, and ANet-QA \cite{caba2015activitynet}. We empirically observe that the default version of GPT-3.5-Turbo would significantly impact evaluation performance. Thus, we also report the possible GPT-3.5 versions for evaluation.}
    \vspace{-0.9em}
\resizebox{0.48\textwidth}{!}{
        \begin{tabular}{l|c|cccccc}
        \hline
        \multirow{2}{*}{Methods}  & \multirow{2}{*}{Size} & \multicolumn{2}{c}{MSVD-QA} & \multicolumn{2}{c}{MSRVTT-QA} & \multicolumn{2}{c}{ANet-QA}  \\
         &  & Acc & Score & Acc & Score & Acc & Score  \\ \hline
         VideoLLaMA \cite{zhang2023video} & 7B &  51.6 & 2.5 & 29.6 & 1.8 & 12.4 & 1.1  \\ 
         LLaMA-Adapter~\cite{zhang2023llama} & 7B & 54.9 & 3.1 &43.8 &2.7 & 34.2 & 2.7  \\ 
         VideoChat~\cite{li2023videochat} & 7B & 56.3 & 2.8 & 45.0 & 2.5 & 26.5 & 2.2  \\ 
         VideoChatGPT~\cite{maaz2023video} & 7B & 64.9 & 3.3 & 49.3 & 2.8 & 35.2 & 2.7  \\ 
         BT-Adapter~\pub{CVPR 24}~\cite{liu2023one} & 7B & 67.5 & 3.7 & 57.0 & 3.2 & 45.7 & 3.2   \\
         Chat-UniVi~\pub{CVPR 24}~\cite{jin2023chat} & 7B  & 65.0 & 3.6 & 54.6 & 3.1 & 45.8 & 3.2   \\
         LLaMA-VID~\pub{ECCV 24}~\cite{li2023llama} & 7B & 69.7 & 3.7 & 57.7 & 3.2 & 47.4 & 3.3   \\
         LLaMA-VID~\pub{ECCV 24}~\cite{li2023llama} & 13B & 70.0 & 3.7 & 58.9 & 3.3 & 47.5 & 3.3   \\
         VideoChat2~\pub{ECCV 24}~\cite{li2023mvbench} & 7B & 70.0 & 3.9 & 54.1 & 3.3 & 49.1 & 3.3  \\
         ST-LLM~\pub{ECCV 24}\cite{DBLP:journals/corr/abs-2404-00308} & 7B & \cellcolor{green!20}74.6 & \cellcolor{green!20}3.9 & \cellcolor{green!20}63.2 & \cellcolor{green!20}3.4 & \cellcolor{green!20}50.9 & 3.3  \\ 
         \hline
          DynFocus (Turbo-16k) & 7B & 72.3 & 3.9 & 59.8 & 3.4 & 49.4 & \cellcolor{green!20}3.4 \\ 
          DynFocus (Turbo-0613) & 7B & \cellcolor{green!20}74.8 & \cellcolor{green!20}4.0 & \cellcolor{green!20}62.8 & \cellcolor{green!20}3.6 & \cellcolor{green!20}50.3 & \cellcolor{green!20}3.4 \\ 
        \hline
        \end{tabular}
}
 \vspace{-0.8em}
\label{tab:qa}
\end{table}

\begin{table}[t]
\setlength{\tabcolsep}{3pt}
\centering
\caption{Performance comparisons on VCG-Bench. $\dagger$ represents the
version that first fine-tuned on all the dataset, and further post-tuning on VideoChatGPT-100K~\cite{maaz2023video} with a smaller learning rate.}
 \vspace{-0.7em}
\resizebox{0.48\textwidth}{!}{
\begin{tabular}{l|c|ccccc|c}
\hline
  Methods & Size & CI &  DO & CU & TU & CO & Avg. \\ 
  \hline
 VideoLLaMA \cite{zhang2023video}   & 7B & 1.96 & 2.18 & 2.16 & 1.82  & 1.79 &  1.98 \\
 LLaMA-Adapter~\pub{CVPR 23}~\cite{zhang2023llama} & 7B & 2.03 & 2.32 & 2.30 & 1.98  & 2.15 & 2.16 \\
 VideoChat \cite{li2023videochat}  & 7B & 2.23 & 2.50 & 2.53 & 1.94  & 2.24 & 2.29 \\
 VideoChatGPT~\pub{ACL 24} \cite{maaz2023video}  & 7B & 2.40 & 2.52 & 2.62 & 1.98  & 2.37 & 2.38 \\
 BT-Adapter~\pub{CVPR 24}~\cite{liu2023one}     & 7B & 2.68 & 2.69 & 3.27 & 2.34  & 2.46 & 2.69 \\
 VTimeLLM \cite{huang2023vtimellm} & 7B & 2.78 & \cellcolor{green!20}3.10 & 3.40 & 2.49  & 2.47 & 2.85 \\
 Chat-UniVi~\pub{CVPR 24}~\cite{jin2023chat}  & 7B & 2.89 & 2.91 & 3.46 & \cellcolor{green!20}2.89 & \cellcolor{green!20}2.81 & 2.99 \\
 LLaMA-VID~\pub{ECCV 24}~\cite{li2023llama}  & 7B & 2.96 & 3.00 & 3.53 & 2.46  & 2.51 & 2.89 \\
 VideoChat2~\pub{CVPR 24} \cite{li2023mvbench}  & 7B & 3.02 & 2.88 & 3.51 & 2.66  & \cellcolor{green!20}2.81 & 2.98 \\
 PLLaVA~\pub{CVPR 24} \cite{li2023mvbench}  & 7B & 3.21 & 2.86 & 3.62 & 2.33 & \cellcolor{green!20}2.93 & 3.12 \\
 ST-LLM~\pub{ECCV 24}~\cite{DBLP:journals/corr/abs-2404-00308} & 7B & \cellcolor{green!20}3.23 & 3.05 & \cellcolor{green!20}3.74 & \cellcolor{green!20}2.93  & \cellcolor{green!20}2.81 & \cellcolor{green!20}3.15 \\
 \hline
 DynFocus & 7B& 3.12 & 3.11 & 3.68  & 2.57 & 2.74 & 3.05 \\
 DynFocus$^\dag$ & 7B & \cellcolor{green!20}3.27 & \cellcolor{green!20}3.15 & \cellcolor{green!20}3.78  & 2.86 & 2.78 & \cellcolor{green!20}3.17\\
\hline
\end{tabular}
}
\vspace{-1em}
\label{tab:VCG-Bench}
\end{table}

\begin{table}[ht]
\setlength{\tabcolsep}{3pt}
\centering
\caption{Performance Comparisons on LV-Bench. Input shows the number of frames each model actually process when testing. $\dagger$ denotes the optimal results when adopting different number of $L$ on 200 input video frames.}
\vspace{-0.8em}
\resizebox{0.48\textwidth}{!}{
       \begin{tabular}{l|c|c|cccccccc}
        \hline
        \textbf{Method} &  
        \textbf{Size} & \textbf{Input} & \textbf{ER} &  \textbf{EU} &  \textbf{KIR} &  \textbf{TG} &  \textbf{Rea} &  \textbf{Sum} &  \textbf{Overall} \\
        \hline
        \multicolumn{10}{c}{\textbf{\textit{Short Video MLLMs}}} \\
        \hline
        TimeChat~\pub{CVPR 24}~\cite{Ren2023TimeChat} & 7B & 96 f & 21.9 & 21.7 & 25.9 & 22.7 & 25.0 & 24.1 & 22.3 \\
        \textcolor{gray}{PLLaVA~\cite{xu2024pllava}} & \textcolor{gray}{34B} & \textcolor{gray}{16 f} & \textcolor{gray}{25.0} & \textcolor{gray}{24.9} & \textcolor{gray}{26.2} & \textcolor{gray}{21.4} & \textcolor{gray}{30.0} & \textcolor{gray}{25.9} & \textcolor{gray}{26.1} \\
        \textcolor{gray}{LLaVA-NeXT~\cite{zhang2024llavanextvideo}} & \textcolor{gray}{34B} & \textcolor{gray}{32 f} & \textcolor{gray}{30.1} & \textcolor{gray}{31.2} & \textcolor{gray}{34.1} & \textcolor{gray}{31.4} & \textcolor{gray}{35.0} & \textcolor{gray}{27.6} & \textcolor{gray}{32.2} \\
        \textcolor{gray}{GPT-4o~\cite{openai2024gpt4o}} & - & \textcolor{gray}{10 f} & \textcolor{gray}{26.5} & \textcolor{gray}{23.7} & \textcolor{gray}{28.3} & \textcolor{gray}{21.4} & \textcolor{gray}{28.0} & \textcolor{gray}{32.8} & \textcolor{gray}{27.0} \\
        \hline
        \multicolumn{10}{c}{\textbf{\textit{Long Video MLLMs}}} \\
        \hline
        MovieChat~\pub{CVPR 24}~\cite{song2023moviechat} & 7B & $\sim$10k f & 21.3 & 23.1 & 25.9 & 22.3 & 24.0 & 17.2 & 22.5 \\
        LLaMA-VID~\pub{ECCV 24}~\cite{li2023llama} & 13B & $\sim$10k f & 25.4 & 21.7 & 23.4 & 26.4 & 26.5 & 17.2 & 23.9 \\
        LWM~\cite{Liu2024WorldMO} & 7B & $\sim$4k f & 24.7 & 24.8 & 26.5 & 28.6 & 30.5 & 22.4 & 25.5 \\
        \textcolor{gray}{Gemini 1.5 Pro~\cite{reid2024gemini}} & - & \textcolor{gray}{$\sim$4k f } & \textcolor{gray}{32.1} & \textcolor{gray}{30.9} & \textcolor{gray}{39.3} & \textcolor{gray}{31.8} & \textcolor{gray}{27.0} & \textcolor{gray}{32.8} & \textcolor{gray}{33.1} \\
        \hline
        DynFocus$^{\dag}$ ($L=25, K/L=0.8$) & 7B & 200 f & 27.9 & 30.3 & 31.2 & 25.4 & 31.8 & 32.8 & 30.4 \\
        DynFocus$^{\dag}$ ($L=50, K/L=0.8$)  & 7B & 200 f & 28.6 & 31.8 & 32.6 & 27.2 & 35.3 & \cellcolor{green!20}34.4 & 31.8 \\
        DynFocus$^{\dag}$ ($L=60, K/L=0.6$)  & 7B & 200 f & 29.9 & \cellcolor{green!20}33.7 & \cellcolor{green!20}35.1 & 25.5 & 33.3 & 26.2 & 32.6 \\
        DynFocus$^{\dag}$ ($L=70, K/L=0.4$) & 7B & 200 f & \cellcolor{green!20}31.8 & 33.5 & 32.6 & \cellcolor{green!20}28.7 & \cellcolor{green!20}34.8 & 31.3 & \cellcolor{green!20}32.9 \\
        DynFocus$^{\dag}$ ($L=80, K/L=0.4$) & 7B & 200 f & 31.1 & 33.5 & 31.6 & 28.6 & 33.8 & 24.1 & 31.8 \\
        \hline
    \end{tabular}
}
\label{tab:LV-Bench}
\vspace{-1em}
\end{table}

\begin{table*}[t]
\small
\centering
\caption{The overall performances on MLVU. Two input strategies are adopted in evaluation: Uniform Sampling (\textit{N fr}), which evenly samples N frames from the video; Frame Rate Sampling (\textit{N fps}), which samples N frames per second. $\dag$ denotes proprietary models.} 
\resizebox{0.9\textwidth}{!}{
\begin{tabular}{l|c|ccccccccccc}
\hline
\multirow{2}{*}{\textbf{Methods}} & \multirow{2}{*}{\textbf{Input}}  & \multicolumn{3}{c}{\textbf{Holistic}} & \multicolumn{4}{c}{\textbf{Single Detail}}    & \multicolumn{2}{c}{\textbf{Multi Detail}} & \multirow{2}{*}{\textbf{M-Avg}} & \multirow{2}{*}{\textbf{G-Avg}} \\ 
&~& TR   &AR & VS  & NQA &ER &PQA
& SSC   & AO &AC \\     
        \hline
         \multicolumn{13}{c}{\textit{\textbf{Short Video MLLMs}}} \\
         \hline
         VideoChat~\cite{li2023videochat}& 16 f  & 33.0& 32.0 & 2.31 & 27.0 & 32.1 & 27.6  & 5.01 & 24.3 &28.6 &29.2 &3.66  \\
        Video-ChatGPT~\pub{ACL 24}~\cite{maaz2023video}& 100 f & 26.9 & 24.0 & 2.31 & 40.3 & 42.0 & 29.9  & 5.48 & 25.1 &31.1 &31.3 &3.90  \\
        Video-LLaMA2~\cite{DBLP:journals/corr/abs-2406-07476}& 16 f  & 54.5  &41.5  &2.34  &39.4  &33.5  &35.4   & 5.22  &18.5  &25.7 &35.5 &3.78  \\
         VideoChat2~\pub{CVPR 24} \cite{li2023mvbench} & 16 f & 74.6 & 51.5 & 2.57 & 42.0 &  47.4 & 43.8  & 5.04 & 22.8 &29.6 &44.5 &3.81  \\
          Video-LLaVA~\cite{DBLP:conf/emnlp/LinYZCNJ024}& 8 f   &71.6  & 57.0  &2.43  & 53.2 &45.2  & 48.4   &5.25  &20.1  & 35.9 &47.3 &3.84  \\
        \hline
        \multicolumn{13}{c}{\textit{\textbf{Long Video MLLMs}}}\\
        \hline
         MovieChat~\pub{CVPR 24}~\cite{song2023moviechat} & 2048 f   & 29.5  & 25.0  &2.33  &24.2  &24.7 &25.8  &3.23    &\cellcolor{green!20}28.6  &22.8 &25.8 &2.78    \\
         Movie-LLM~\cite{DBLP:journals/corr/abs-2403-01422} & 1 fps  & 30.0  & 29.0  &2.88  &29.6  &24.7  &24.1    &5.00  &20.5 &24.8 &26.1 &3.94   \\
         TimeChat~\pub{CVPR 24}~\cite{Ren2023TimeChat} & 96 f  & 23.1  & 27.0  &2.54  &24.5  &28.4  &25.8   &4.29  &24.7  &32.0 &30.9 &3.42  \\
         LLaMA-VID~\pub{ECCV 24}~\cite{li2023llama} & 1 fps  & 50.8  & 34.5  &3.22  &30.1  &32.7  &32.5    &5.22  &23.9 &27.8 &33.2 &4.22   \\
        MA-LMM~\pub{CVPR 24}~\cite{he2024malmm}& 1000 f & 51.9  & 35.5  &2.12  &43.1  &38.9  &35.8    &4.80  &25.1 &24.3 &36.4 &3.46   \\
        MiniGPT4-Video~\cite{DBLP:journals/corr/abs-2404-03413} & 90 f & 70.9  & 52.5  &2.64  &49.0  &\cellcolor{green!20}48.6  &44.5   &4.07  &23.2 &23.0 &44.5 &3.36   \\
        \hline
        DynFocus ($L=25, K/L=0.8$)  & 16 f  & \cellcolor{green!20}75.4  & \cellcolor{green!20}60.5  & \cellcolor{green!20}3.36 & \cellcolor{green!20}50.6  & \cellcolor{green!20}42.3  & \cellcolor{green!20}50.5   & \cellcolor{green!20}5.34  & 26.2 & \cellcolor{green!20}32.6 & \cellcolor{green!20}48.3 & \cellcolor{green!20}4.35   \\
        DynFocus ($L=25, K/L=0.8$)   & 32 f  & \cellcolor{green!20}76.2  & \cellcolor{green!20}60.9 & \cellcolor{green!20}3.36 & \cellcolor{green!20}55.5 & 41.5  & \cellcolor{green!20}54.0   & \cellcolor{green!20}5.39  & \cellcolor{green!20}26.8 & \cellcolor{green!20}32.8 &\cellcolor{green!20}49.6 & \cellcolor{green!20}4.38   \\
        \hline
        \textcolor{gray}{GPT-4o$^\dag$~\cite{openai2024gpt4o}} & \textcolor{gray}{0.5 fps}  & \textcolor{gray}{87.4}  & \textcolor{gray}{74.5}  & \textcolor{gray}{4.90}  & \textcolor{gray}{64.8}  & \textcolor{gray}{57.1}  & \textcolor{gray}{65.1}  & \textcolor{gray}{6.69}  & \textcolor{gray}{56.7} & \textcolor{gray}{46.3} & \textcolor{gray}{64.6} & \textcolor{gray}{5.80}  \\
\hline
 \end{tabular}
}
 \vspace{-1em}
\label{tab:mlvu}
\end{table*}

\subsection{Evaluation on Short Video Understanding}

\noindent\textbf{Zero-shot Video-question Answering Performance.} In Table~\ref{tab:qa}, we report the results of our \textit{DynFocus} against a bunch of SOTA methods on three widely-used QA benchmarks: MSVD-QA~\cite{DBLP:conf/acl/ChenD11}, MSRVTT-QA~\cite{DBLP:conf/cvpr/XuMYR16}, and ANet-QA~\cite{DBLP:conf/cvpr/HeilbronEGN15}. 
On MSRVTT-QA and MSVD-QA, our model achieves comparable results than published SOTA ST-LLM~\cite{DBLP:journals/corr/abs-2404-00308}. For slightly longer video ANet-QA, our method achieves competitive performance using $\sim$25\% fewer tokens than ST-LLM, exhibiting a balance between accuracy and memory efficiency. Beyond that, we empirically observe that the marginal performance gain on short video dataset gradually decreases as the dataset scale expands during instruction tuning, which can be found in the supplementary material. 

\noindent\textbf{VCG-Bench Performance.}
Table~\ref{tab:VCG-Bench} presents the results on VideoChatGPT~\cite{maaz2023video} in terms of Correctness of Information (CI), Detailed Orientation (DO), Contextual Understanding (CU), Temporal Understanding (TU) and Consistency (CO). Our \textit{DynFocus} outperforms existing video MLLMs on CI, DO, and CU. Notably, it substantially surpasses VideoChat2~\cite{li2023mvbench} on CI despite using fewer instructional dataset. This may be attributed to our DPE module, which supports dynamically mitigating the visual nuisance that could hamper factual correctness. ST-LLM shows slight advantages over ours on TU for two possible reasons: (1) it performs the feature alignment between masked input and unmasked video input, which explicitly emphasizes the temporal relationship. (2) The retained tokens of each frame in ST-LLM is more than ours, and more visual details could compensate for temporal clues when handling short videos.

\begin{table*}[ht]
\centering
\caption{Comparisons on VideoMME with short, medium, and long durations, under the settings of ``without subtitles'' and ``with subtitles''. Notably, our method adopts 224$^{2}$ frame resolution instead of using original resolution. $\dagger$ denotes the model with DPO tuning.}
 \vspace{-0.7em}
\resizebox{0.9\textwidth}{!}{
\begin{tabular}{l|c|c|cccccccc} 
\hline
\multicolumn{1}{c|}{\multirow{2}{*}{\textbf{Models}}} & \multirow{2}{*}{\textbf{Input}} & \multirow{2}{*}{\begin{tabular}[c]{@{}c@{}}\textbf{LLM}\\\textbf{Size}~~\end{tabular}} & \multicolumn{2}{c}{\textbf{Short (\%)}} & \multicolumn{2}{c}{\textbf{Medium (\%)}} & \multicolumn{2}{c}{\textbf{Long (\%)}} & \multicolumn{2}{c}{\textbf{Overall (\%)}}  \\ 
&   &   & w/o subs      & w/ subs     & w/o subs      & w/ subs    & w/o subs      & w/ subs                & w/o subs      & w/ subs                    \\ 
\hline

LLaMA-VID~\pub{ECCV 24}~\citep{li2023llama}   & 1 fps & 7B  & - & -  & -  & - & - & - & 25.9 & - \\
Video-LLaVA~\pub{EMNLP 24}~\citep{lin2023video}   & 8 f & 7B  & 45.3 & 46.1  & 38.0  & 40.7 & 36.2 & 38.1 & 39.9 & 41.6                       \\
ST-LLM~\pub{ECCV 24}~\citep{liu2023one}  & 16 f & 7B  & 45.7  & 48.4  & 36.8   & 41.4   & 31.3  & 36.9  & 37.9  & 42.3   \\
VideoChat2~\pub{CVPR 24}\citep{li2023mvbench}  & 16 f & 7B  & \cellcolor{green!20}48.3   & \cellcolor{green!20}52.8   & 37.0   & 39.4 & 33.2   & 39.2   & 39.5  & 43.8 \\
Chat-UniVi~~\pub{CVPR 24}~\citep{jin2023chat}  & - & 7B   & 45.7   & 51.2  & \cellcolor{green!20}40.3  & \cellcolor{green!20}44.6  & \cellcolor{green!20}35.8 & \cellcolor{green!20}41.8  & \cellcolor{green!20}40.6  & \cellcolor{green!20}45.9 \\
DynFocus ($L=25, K/L=0.8$) & 16 f & 7B   & \cellcolor{green!20}50.9   & \cellcolor{green!20}53.7  & \cellcolor{green!20}43.7  & \cellcolor{green!20}46.0  & \cellcolor{green!20}37.7 & \cellcolor{green!20}43.6  & \cellcolor{green!20}44.1  & \cellcolor{green!20}47.8 \\
\textcolor{gray}{LLaVA-NeXT$^{\dag}$~\citep{zhang2024llavanextvideo}} & - & \textcolor{gray}{34B}   & \textcolor{gray}{61.7}   & \textcolor{gray}{65.1}  & \textcolor{gray}{50.1}  & \textcolor{gray}{52.2}  & \textcolor{gray}{44.3}  & \textcolor{gray}{47.2}   &  \textcolor{gray}{52.0}  & \textcolor{gray}{54.9}                       \\
\textcolor{gray}{VILA-1.5~\citep{lin2023vila}}                 &       -        &  \textcolor{gray}{34B} & \textcolor{gray}{68.1} & \textcolor{gray}{68.9}  & \textcolor{gray}{58.1} & \textcolor{gray}{57.4}            & \textcolor{gray}{50.8} & \textcolor{gray}{52.0}         & \textcolor{gray}{59.0} & \textcolor{gray}{59.4}              \\ 
\hline
 \end{tabular}}
\label{tab:videomme}
\end{table*}

\subsection{Evaluation on Long Video Understanding}
To demonstrate the advantage of our dynamic cooperative setting, we conduct experiment on three newly released long-term video benchmark. The detailed description for each benchmark are elaborated in supplementary material. 
\noindent\textbf{MLVU-Bench performance.} The performance of individual task and the average performance of multi-choice task (M-Avg, within 0-100\%) and generation task (G-Avg, within 0.0-10.0) are both reported in Table~\ref{tab:mlvu}. We have following observations: (1) our \textit{DynFocus} surpasses all the open-sourced video MLLMs with a clear-cut performance gain on M-Avg and G-Avg, and it nearly consistently ranks top-2 position on individual tasks. (2) For TR, AR, and VS tasks that require an thorough understanding of entire video, our method achieves the best. We attribute this to our dynamic cooperative network's ability to balance intricate spatial details with broader temporal perception without introducing external visual nuisance. (3) Most approaches find AO and AC tasks challenging due to their sensitivity to the temporal clues, which requires recalling multiple nuanced details from lengthy videos. Although not being further fine-tuned on long-term video dataset like MovieChat, our model still performs competitively. (4) However, our method struggles with ER task that needs ego-based perspectives, likely due to the requirement for ego-centric dataset like EgoQA~\cite{egoqa} in VideoChat2.
 
\noindent\textbf{LV-Bench Performance.}
We assess six core capabilities of our model on LV-Bench: Temporal Grounding (TG), Summarization (Sum), Reasoning (Rea), Entity Recognition (ER), Event Understanding (EU), and Key Information Retrieval (KIR). The average duration of each video exceeds \textbf{1 hour}. Following~\cite{DBLP:journals/corr/abs-2406-08035}, we select several publicly evaluated methods as baselines, with results shown in Table~\ref{tab:LV-Bench}. Interestingly, some methods that excel on short videos perform almost randomly in answer selection. Remarkably, our \textit{DynFocus} achieves the best of 32.9\% among all the open-sourced 7B models, even outperforming PLLaVA~\cite{li2023mvbench} with 34B parameters.




\begin{figure}[ht]
\vspace{-1em} 
\centering
\hspace{-1em} 
\includegraphics[scale=0.16]{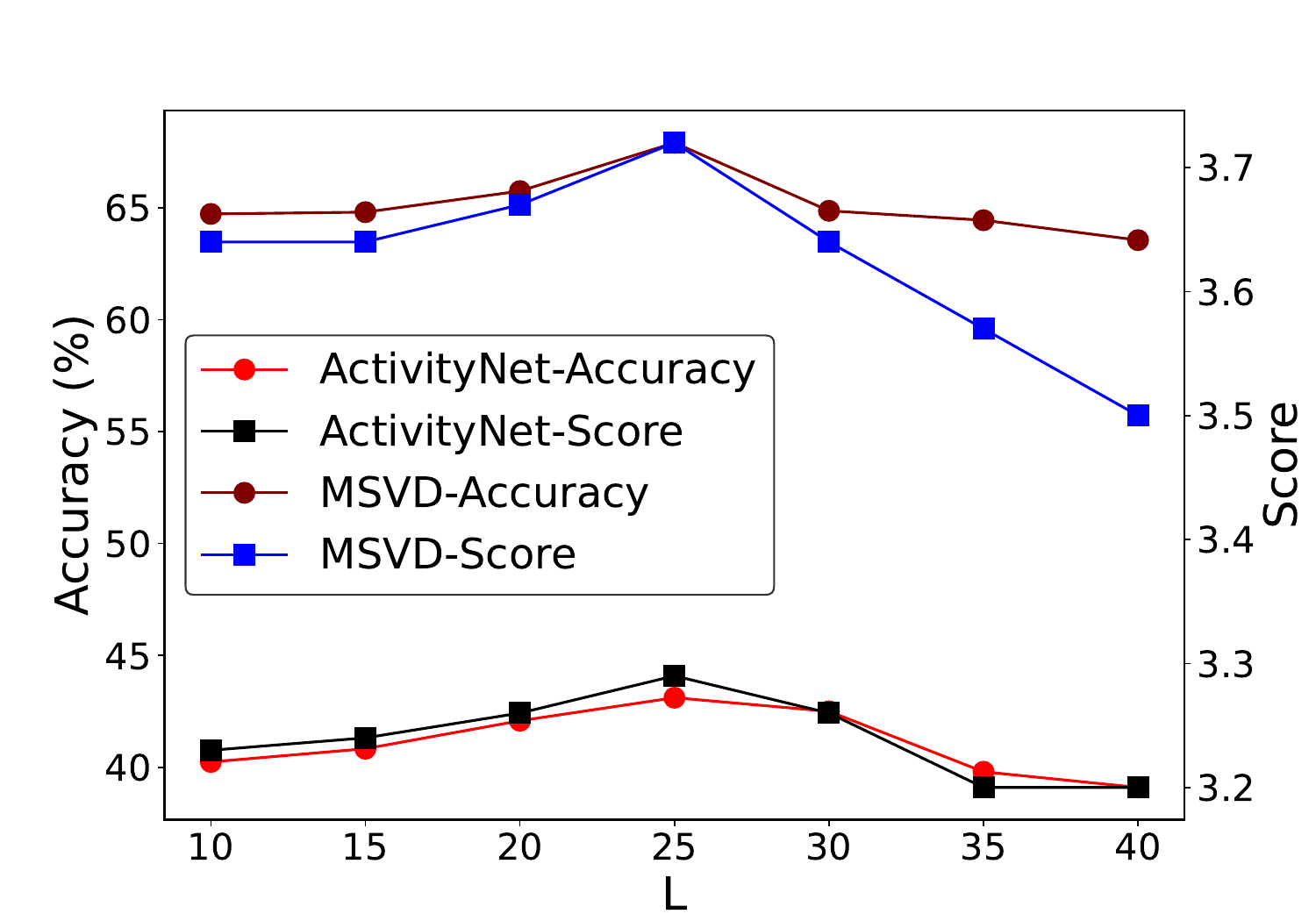}
\hspace{-0.25cm}
\includegraphics[scale=0.16]{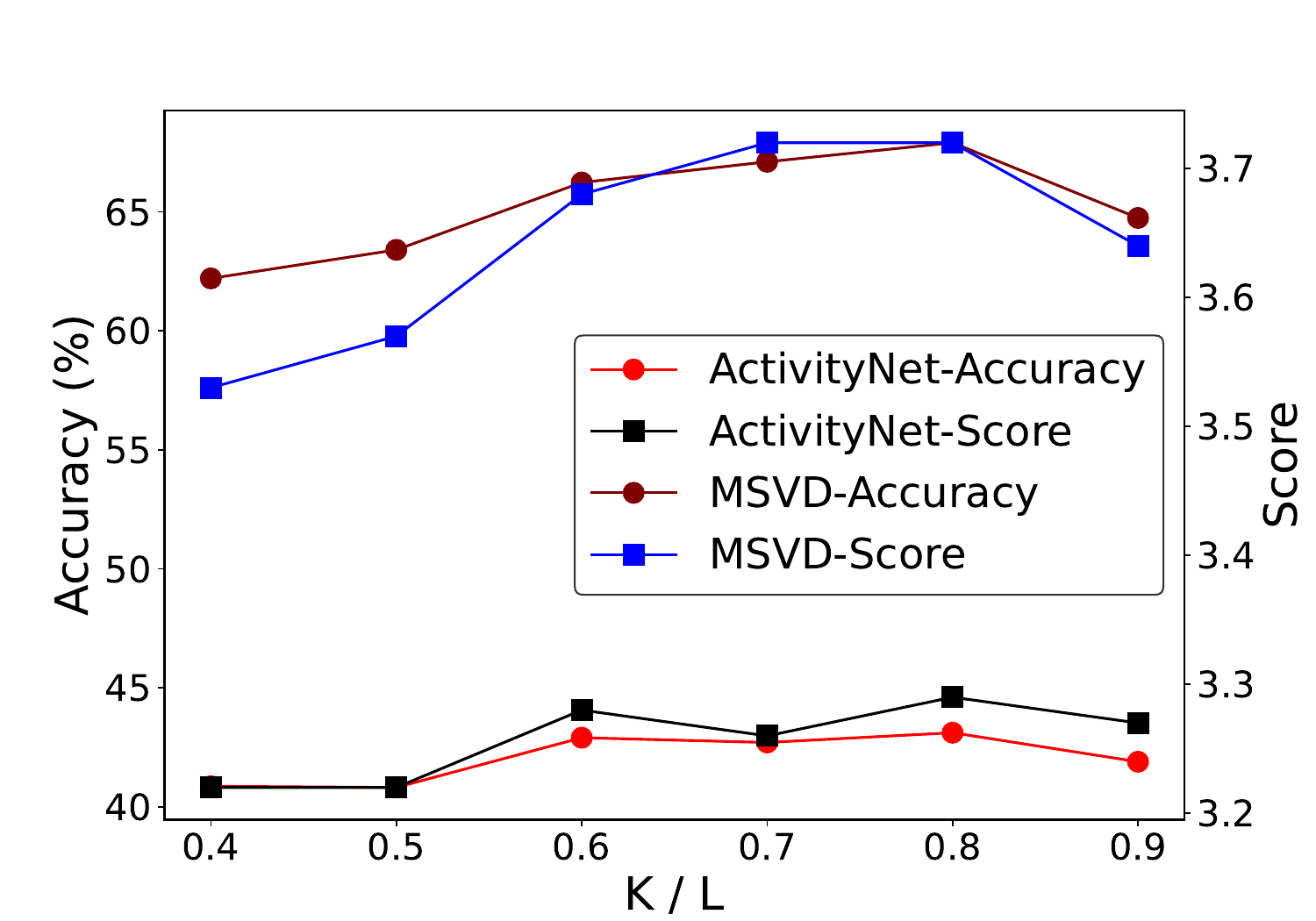}

\vspace{-0.1cm} 
\caption{(a) and (b) illustrate the performance with different number of event prototypes and different ratio of filtered event prototypes, respectively.}
\label{fig:DLI}
\vspace{-1.4em}
\end{figure}

\noindent\textbf{VideoMME Benchmark Performance.}
VideoMME benchmark spans across three kinds of durations, VideoMME-S ($\sim$1.3min), VideoMME-M ($\sim$8.5min), and VideoMME-L ($\sim$0.7h). Most videos include both subtitles and audios, which helps us investigate the performance gain from additional information sources. Table~\ref{tab:videomme} compares our results with other representative video MLLMs. Notably, our \textit{DynFocus} consistently achieves impressive advantage across different lengths of video with subtitles and without subtitles. Specifically, it exhibits the overwhelming advantage over SOTA ST-LLM and VideoChat2. Remarkably, the version of \textit{DynFocus} without subtitles reaches an overall accuracy of 44.1\%, still surpassing ST-LLM with subtitles by 1.8\%. 

\subsection{Evaluation on Video Hallucination} 
Our method also achieves the competitive performance on addressing video hallucination on VideoHallucer~\cite{DBLP:journals/corr/abs-2406-16338}. We report the detailed results and give further analysis in the supplementary materials due to the space limitation. 
\subsection{Component-wise Analysis}

\begin{table}[t]
\centering                         
\caption{We report the results using different numbers of token to encode the frame with $b_{t}=0$ and $b_{t}=1$. Specifically, 40 tokens involves 22 multi-grained prototypes, i.e., $\mathbf{G}_{t}$, 16 tokens in each filtered event prototype $\mathbf{h}_{t}$, 1 global content token, and 1 text-guided token. 256 represents the original number of tokens without compression. $\left| \cdot \right|$ denotes the token number.}
\vspace{-0.5em}
\footnotesize                      
\scalebox{0.87}{
\begin{tabular}{cc|cc|cc|c}
  \hline
  \multirow{2}{*}{$\left|\mathbf{U}_{b_{t}=0} \right|$} & \multirow{2}{*}{$\left|\mathbf{U}_{b_{t}=1}\right|$} & \multicolumn{2}{c|}{\textbf{MSVD-QA}} & \multicolumn{2}{c|}{\textbf{ANet-QA}} & \textbf{VCG-Bench}   \\
     & & Acc & Score & Acc & Score & Score \\ 
  \hline
  0 & 40 & 63.7 & 3.5 & 41.4 & 3.2 & 2.57  \\
  0 & 256 & 65.6 & 3.5 & 42.1 & 3.2 & 2.65  \\
  2 & 256 & 68.4  & 3.7 & 44.3 & 3.4 & 2.85  \\
  2 & 2 & 62.0 & 3.5 & 40.5 & 3.2 & 2.38   \\
  2 & 0 & 58.2 & 3.3 & 38.6 & 2.9 & 2.21  \\
  2 & 40 & 67.9 & 3.7 & 43.1 & 3.3 & 2.81  \\
  \hline
\end{tabular}}
\label{tab:cooperation}
\end{table}

 \begin{figure}
    \centering
    \vspace{-1em}
    \includegraphics[width=0.9\linewidth]{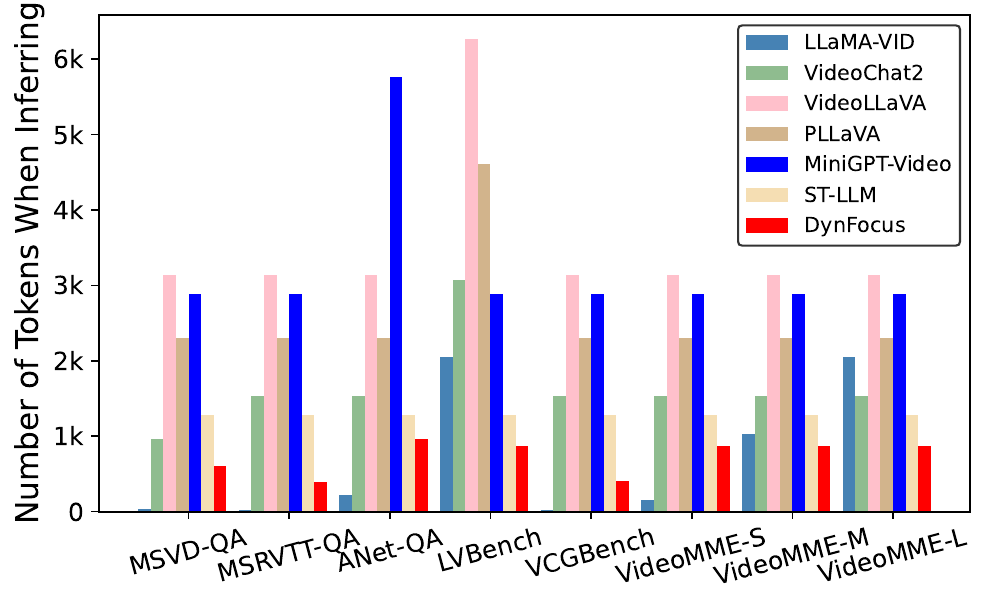}
     \vspace{-1em}
    \caption{Token number comparison with different methods on different benchmark datasets. We calculate their token number using their released code snippet regarding loading video.}
    \label{fig:token_number}
     \vspace{-1em}
\end{figure}

\begin{figure}
    \centering
    \includegraphics[width=1.04\linewidth]{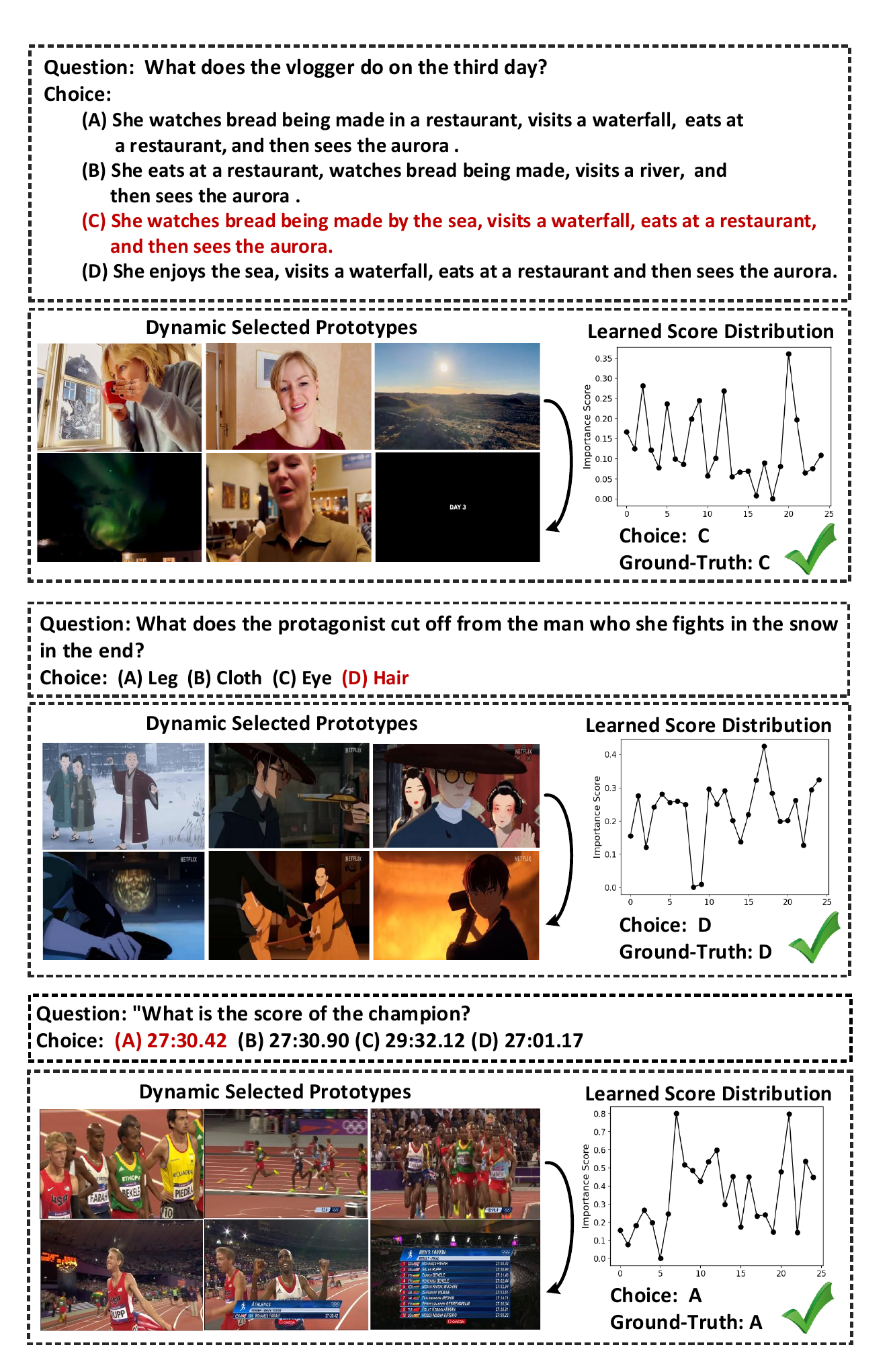}
    \vspace{-2em}
    \caption{We showcase the filtered event prototypes focused by DPE module on LV-Bench. To save space, we only showcase the prototype with top-6 score sequentially. The figure at the right-bottom corner illustrates the learned score distribution on the event prototype candidates ($L$=25) obtained by DPC-KNN.}
    \label{fig:focus}
    \vspace{-1.5em}
\end{figure}

\noindent\textbf{The Effect of Number of Initial Event Prototypes.} As depicted in Figure~\ref{fig:DLI}a, we observe that increasing $L$ (10 $\sim$ 25) brings a consistent gain in the overall accuracy. It indicates that sufficient number of prototypes could divide the video into more fine-grained events, which would offer more abundant visual clues for accurate question answering. However, when $L$ increases greater than 25, the performance begins to drop. This phenomenon can be explained by that increasing prototypes would hamper the intrinsic temporal structure as well as consistency. More results towards long-term video can be referred in the supplementary materials.

\noindent\textbf{The Effect of Dynamic Selection.} As shown in Figure~\ref{fig:DLI}b, a similar pattern can be observed by varying the ratio $K/L$, where $L$ and $K$ represent the number of event prototype candidates obtained by DPC-KNN clustering and filtered ones, respectively. The smaller ratio of filtered prototypes may be not enough to cover all useful visual clues, whereas the larger ratio still maintain much non-essential visual nuisance, thereby disturbing content understanding.

\noindent\textbf{Visualization of Focused Frame by DPE module.} As explained in DPE module, the higher score in Figure~\ref{fig:focus} indicates greater contribution to question answering. Taking the first case as an example, the question asks about the occurred events and their corresponding sequence. Although these frames with great contribution encoded by $\textit{Cones}$ could provide detailed visual semantics like sea, waterfall, and aurora, they may not offer sufficient temporal clues to determine the order of events. This information can be encapsulated by its complementary part, akin to $\textit{Rods}$, which provides broader receptive field for capturing motion.

\noindent\textbf{The Effect of Cooperation between $\textit{Cones}$ and $\textit{Rods}$.} We introduce several variants to validate the benefit of cooperation between $\textit{Cones}$ and $\textit{Rods}$. As reported in Table~\ref{tab:cooperation}, $\left|\mathbf{U}_{b_{t}=0} \right| = 0$ indicates that we discard the tokens encoded by $\textit{Rods}$. We observe significant performance drop 1.7\% on MSVD-QA compared with our full model. The similar pattern can be observed when completely dropping the tokens encoded by $\textit{Cones}$. Moreover, although the model exhibits the best results without token compression for those important frames, it still encounter the scalability issues when extending to long-term videos, struggling to balance memory efficiency and accuracy. Figure~\ref{fig:token_number} shows the comparison of total token usage, demonstrating superior advantages over existing methods.
\vspace{-1em}
\section{Conclusion}
In this paper, we develop a dynamic cooperative network for memory-efficient encoding. We experimentally delve into the network behavior and find that dynamic encoding could simultaneously achieves fine spatial visual appearance understanding and coarse temporal dynamics perception using affordable tokens, striking a balance between answering accuracy and memory efficiency. Our model achieves superior performance with substantially few tokens on both short and long video benchmarks. Moreover, our model also demonstrates the great potential on addressing video hallucination. 

\section{Acknowledgment}
This work was supported by the National Natural Science Foundation of China (62302045) and the Beijing Institute of Technology Special-Zone.

{\small
\bibliographystyle{ieee}
\bibliography{cvpr}
}
\clearpage

\appendix
\section*{Abstract of Appendix} This appendix provides the implementation of redundancy estimation (Appendix~\ref{redundancy}), additional discussions (Appendix~\ref{more_discussions}), more implementation details (Appendix~\ref{more_imp}), more visualization results (Appendix~\ref{more_vis}), and case study (Appendix~\ref{case}).

\section{Redundancy Estimation}
\label{redundancy}
In this section, we provide the details about how to estimate the ratio of temporal repetitive frames and answer-irrelevant frames, which is denoted as $r_{d}$ and $r_{a}$, respectively. Specifically, given a $T$-frame video, we use CLIP-ViT~\cite{DBLP:conf/icml/RadfordKHRGASAM21} to extract the representation for each video frame and its text part.

For temporal repetitive frames, we calculate the cosine similarities of features between consecutive frames, denoted as $s_{t} = \mathrm{cos}(\mathbf{f}_{t}, \mathbf{f}_{t+1})$. We then collect all scores into a score vector $\mathbf{s} = \{s_{t}\}_{t=1}^{T-1}$ and apply min-max normalization. This process can be summarized as,
\begin{align}
 r_{d} = \frac{\sum_{t=1}^{T-1} \mathbf{I}(s_{t}>0.6)}{T-1},
 \label{eqn:rd}
\end{align}
where $\mathbf{I}(\cdot)$ is the indicator function, defined as $\mathbf{I}(\mathbf{x}) = 1$ if $\mathbf{x}$ is true, and $\mathbf{I}(\mathbf{x}) = 0$ if $\mathbf{x}$ is false. 

For answer-irrelevant frames, we compute their similarity using $s_{t} = \mathrm{cos}(\mathbf{f}_{t}, \mathbf{q}||\mathbf{a})$, where $\mathbf{q} || \mathbf{a}$ represents the token-wise concatenation of question and answer feature. After applying min-max normalization, we mark a frame as redundant when its frame-to-text similarity falls below a certain threshold. This is summarized as,
\begin{align}
r_{a} = \frac{\sum_{t=1}^{T}\mathbf{I}(s_{t}<0.4)}{T}.
\label{eqn:ra}
\end{align}
Notably, for each benchmark dataset, we randomly sample 20 videos to calculate the average value of redundancy ratio $r_{d}$ and $r_{a}$ as a rough redundancy estimation. 

\section{Additional Discussions}
\label{more_discussions}
\subsection{Component-wise Training State on Model Performance}
We conduct the extensive experiment to explore the effect of different components with different training state. As can be seen in Table~\ref{tab:frozen}, only locking LLM and DPE+CCE module in the first stage exhibits the best, which achieves a obvious performance gain of 0.16 on VCG-Bench. This can be explained that DPE+CCE$^{\dag}$ primarily undertakes the effective feature encoding, whereas the projector $\mathcal{F}_{\text{fine}}, \mathcal{F}_{\text{coarse}}$ may be only responsible for bridging the semantic gap between video content and LLM, respectively. Therefore, the learned knowledge preserved in DPE+CCE$^{\dag}$ in the first stage may not be well adapted to learning of the second stage. In the second stage, unlocking DPE+CCE$^{\dag}$ achieves the substantial performance gain. This may be due to that the knowledge learned in the second stage focuses on video reasoning (for example, which part need to be focused?), which keeps consistent with the design motivation of DPE+CCE$^{\dag}$. 

\begin{table}[ht]
\setlength{\tabcolsep}{3pt}
\centering
\vspace{-0.8em}
    \resizebox{0.48\textwidth}{!}{
        \begin{tabular}{ccc|ccc|cc|c}
        \hline
        \multicolumn{3}{c|}{Vision-Language Alignment} & \multicolumn{3}{c|}{Instruction Tuning}                  & \multicolumn{2}{c|}{\textbf{MSVD-QA}} & \textbf{VCG-Bench}                     \\
            DPE+CCE$^{\dag}$ & $\mathcal{F}_{\text{fine}}, \mathcal{F}_{\text{coarse}}$ & LLM & DPE+CCE$^{\dag}$ & $\mathcal{F}_{\text{fine}}, \mathcal{F}_{\text{coarse}}$ & LLM & Acc & Score & Score \\ \hline
        \faUnlock & \faUnlock & \faLock   & \faUnlock & \faUnlock & \faUnlock   & 65.45          & 3.56   &    2.65        \\
        \faLock & \faLock & \faLock   & \faLock & \faUnlock & \faUnlock         & 61.07          & 3.20    &   2.31     \\
        \faUnlock & \faUnlock & \faLock   & \faLock & \faUnlock & \faUnlock     & 62.21          & 3.34    &  2.38 \\
        \faLock & \faUnlock & \faLock   & \faUnlock & \faUnlock & \faUnlock     & 67.90          & 3.72    &   2.81     \\
        \hline
        \end{tabular}%
    }
    \caption{Performance Comparisons with training state for different components, which is only pretrained and fine-tuned with video dataset. \faLock~indicates parameters are frozen while \faUnlock~denotes the trainable state. DPE+CCE$^{\dag}$ denotes the DPE module and CCE module without $\mathcal{F}_{\text{fine}}, \mathcal{F}_{\text{coarse}}$.}
\label{tab:frozen}
\end{table}

\subsection{Parameter, Runtime and Memory Complexity}

\textbf{Training Time.} 
Table~\ref{tab:train_time} reports the training hours on 8 A100 GPU w/ and w/o the added modules (CCE and DPE). Notably, the model without DPE+CCE refers to that we represents each video frame with two only tokens similar to LLaMA-VID, whereas the model with DPE+CCE additionally generates the finer tokens for important video frames. The increased training time probably comes from the computation time of the extra tokens in LLM backbone, rather than the actual computation time in DPE+CCE module.
\begin{table}[ht]
\setlength{\tabcolsep}{3pt}
\centering
\vspace{-0.8em}
\scalebox{0.87}{
    \begin{tabular}{llll}
\hline
Model       & Stage1 (PT) & Stage2 (SFT) & Total \\ \hline
w/o DPE+CCE & 5.85         & 19.63       & 25.48 \\ \hline
w DPE+CCE   & 7.75         & 25.35       & 33.10 \\ \hline
\end{tabular}}
\vspace{-0.5em}
\caption{Comparison on training hour of methods without DPE+CCE and with DPE+CCE.}
\label{tab:train_time}
\end{table}

\noindent\textbf{Computation Complexity.} 
Table~\ref{tab:computation} reports the inference cost of each added components on LVBench with 1000 input frames on one A100 GPU. 
The calculated event prototypes correspond to T-DPC, the filtered event prototypes correspond to Dyn. Select., multi-grained spatial object prototypes correspond to S-DPC, and Dyn. Enc. corresponds to $\textit{Cones}$ and $\textit{Rods}$ as depicted in CCE. The S-DPC and T-DPC modules do not have trainable parameters. 
\begin{table}[ht]
\setlength{\tabcolsep}{3pt}
\centering
\vspace{-0.8em}
\scalebox{0.87}{
    \begin{tabular}{ccccc}
\hline
\multicolumn{2}{c}{Modules} & \multicolumn{1}{c}{\begin{tabular}[c]{@{}c@{}}Inference\\ GFLOPs \end{tabular}} & \multicolumn{1}{c}{Param. (M)} & \begin{tabular}[c]{@{}c@{}}Inference\\ Latency (ms)  \end{tabular}  \\ \hline
\multicolumn{1}{c}{\multirow{2}{*}{CCE}} & S-DPC            & \multicolumn{1}{c}{0.00}                             & \multicolumn{1}{c}{0.00}   &   50.84      

\\  

\multicolumn{1}{c}{}   &     Dyn. Enc.         & \multicolumn{1}{c}{112.16}                         & \multicolumn{1}{c}{30.31}              &   15.96                     \\ \hline
\multicolumn{1}{c}{\multirow{2}{*}{DPE}} & T-DPC       & \multicolumn{1}{c}{0.00}                             & \multicolumn{1}{c}{0.00}  & 608.28                        \\ 
\multicolumn{1}{c}{}                            & Dyn. Select. & \multicolumn{1}{c}{12.91}                             & \multicolumn{1}{c}{11.87}              &   3.02                     \\ \hline
\end{tabular}}
\caption{Ablative analysis on computation efficiency of added modules.}
\label{tab:CCE}
\vspace{-1em}
\label{tab:computation}
\end{table}

\noindent\textbf{Parameter Budget.} The additional parameter introduced by our designed modules compared with LLaMA-VID are listed in follows:

\textbf{(a) DPE module:} (1) Dynamic Selection (Three MLPs): $\left [ d, \frac{d}{2} \right ] \rightarrow \left [ \frac{d}{2}, \frac{d}{4} \right ] \rightarrow \left [ \frac{d}{4}, 1 \right ]$. 

\textbf{(b) CCE module:} (1) CA module (Two MLPs): $\left [ d, d \right ], \left [ d, d \right ]$; (2) $\mathcal{F}_{\text{coarse}}$ and $\mathcal{F}_{\text{fine}}$ (Two MLPs): $\left [ d, d \right ], \left [ d, d \right ]$

\noindent\textbf{Inference Latency with other baselines.} 
As shown in Table~\ref{tab:metric}, we showcase the comparison of image resolution, averaged inference latency, and input strategies when training. Notably, we achieve the comparable computational efficiency with LLaMA-VID.
\begin{table}[ht]
\setlength{\tabcolsep}{3pt}
\centering
\vspace{-0.8em}
\scalebox{0.7}{
    \begin{tabular}{l|c|ccc|c}
    \hline
    \multirow{2}{*}{Methods}  & \multirow{2}{*}{\textbf{Res.}} & \multicolumn{3}{c|}{\textbf{Inference Latency (s) $\downarrow$}} & \textbf{Training}  \\
     & &  \textbf{MSVD} & \textbf{ANet-QA} & \textbf{VideoMME} & \textbf{Setting} \\ 
    \hline
    LLaMA-VID~\citep{li2023llama}~\pub{ECCV 24} & 224$^2$ & 1.3 & 3.8 & 6.3 & 1 fps \\
    Flash-Vstream~\cite{DBLP:journals/corr/abs-2406-08085} & 224$^2$ & 1.7 & 6.9 & 8.2 & 1 fps \\ 
    DynFocus ($L=25, K/L = 0.8$) & 224$^2$ & 1.4 & 6.4 & 7.8 & 1 fps \\ 
    \hline
    \end{tabular}}
\caption{Comparison on image
resolution, average inference latency, and input strategies when training.}
\label{tab:metric}
\vspace{-1em}
\end{table}

\subsection{Comparison of Method Design with other Methods.}
In this section, we compare the design details with two closely related studies: LLaMA-VID and Chat-Univ. \textbf{(a) Comparison with LLaMA-VID:} LLaMA-VID compresses the each frame into only two tokens: a visual content token and a text-guided context token. Our compression design in $\textit{Rods}$ is somewhat similar to LLaMA-VID. However, the main difference lies in the resolution of input visual signals processed by the text-guided compression module (i.e., $\textit{Context Attention}$). Specifically, LLaMA-VID directly use visual feature at their original resolution. In contrast, our method uses the generated semantic prototypes as the input of $\textit{Rods}$. These prototypes are generated by merging the patch feature with different weight $\rho_{i} \cdot \delta_{i}$, where $i$ denotes the patch index in single frame. (b) \textbf{Comparison with Chat-Univ.} Chat-Univ adopts DPC-KNN clustering algorithm to form clusters both spatially and temporally. Our method differs from Chat-Univ in the following aspects during the clustering process: \textbf{(1) Temporally:} We cluster the frames by calculating the similarity using downsampled features to model more fine-grained temporal relationship, rather than using the feature after global average pooling as in Chat-Univ. This effectively avoids the information loss when performing clustering. \textbf{(2) Spatially:} We use $\mathrm{exp}( \rho_{i} \cdot \delta_{i})$ as weight coefficient when generating the prototype from patch features. \textbf{(3) Token Budget:} The maximum number of tokens per frame in our method is approximately 60\% less than that in Chat-Univ, i.e., 40 tokens versus 112 tokens. \textbf{Essentially}, our model highlights adopting the dynamic encoding, which not only reduces the visual nuisance but also effectively reconciles the spatial details with temporal clues using affordable tokens.
\subsection{Comparison with other Clustering Methods.}
There are multiple clustering algorithm~\cite{macqueen1967some, DBLP:journals/pami/HuangNRL05} available to form the spatial and temporal prototype. To assess the effect of different clustering on model performance, we report the results on two traditional clustering algorithms, $K$-means and weighted $K$-means in Table~\ref{tab:cluster}. To save the time overhead, we train our model using only the video-based dataset.

\begin{table}[t]
\setlength{\tabcolsep}{3pt}
\centering
\vspace{-0.8em}
\scalebox{0.87}{
    \begin{tabular}{l|ccc}
    \hline
    \multirow{2}{*}{Model Variants}  & \multicolumn{2}{c}{\textbf{MSVD-QA}} & \textbf{LV-Bench}  \\
                                     &  Acc & Score & Acc  \\ 
    \hline
    $K$-means~\cite{macqueen1967some} & 66.5 & 3.6 & 23.7 \\
    Weighted $K$-means~\cite{DBLP:journals/pami/HuangNRL05} & 66.8 & 3.6 & 25.1 \\ 
    DPC-KNN & 67.9 & 3.7 & 25.8\\ 
    \hline
    \end{tabular}}
\caption{Effects of different clustering algorithm. }
\label{tab:cluster}
\vspace{-1em}
\end{table}

\begin{table}[t]
\setlength{\tabcolsep}{3pt}
\centering
\scalebox{0.87}{
    \begin{tabular}{l|ccc}
    \hline
    \multirow{2}{*}{Model Variants}  & \multicolumn{2}{c}{\textbf{MSVD-QA}} & \textbf{VCG-Bench}  \\
                                     &  Acc & Score & Score  \\ 
    \hline
    Cross-attention (\textit{Soft}) & 64.74 & 3.61 & 2.56 \\
    Concat. & 66.20 & 3.67 & 2.66 \\ 
    Concat. + Multi-grained & 67.90 & 3.72 & 2.81 \\
    \hline
    \end{tabular}}
    
\caption{Effects of different components in CCE module. Concat. is the concatenation operation.}
\label{tab:CCE}
\vspace{-1em}
\end{table}

\begin{table*}[ht]
\centering
\footnotesize
\caption{\textbf{Ablation of structure and training data.} $\dagger$ represents the results running their official open-sourced code, which adopts the same experimental setting with our \textit{DynFocus}. For fairness, we adopt GPT-3.5-Turbo-16k version for evaluation for all the model in this table.
}
\resizebox{1\textwidth}{!}{
\begin{tabular}{lcc|cc|cc}
\hline
\multirow{3}{*}{Methods} & \multicolumn{1}{c}{\textbf{Vision-Language Alignment}} & \multicolumn{1}{c|}{\textbf{Instruction Tuning}} & \multicolumn{2}{c|}{\multirow{2}{*}{\textbf{MSVD-QA}}} & \multirow{2}{*}{\textbf{VCG-Bench}} & \multirow{2}{*}{\textbf{VideoMME}} \\
\cmidrule(rl){2-2}\cmidrule(rl){3-3}
 &{{Training Datasets}} &{{Training Datasets}} & Acc & Score & Score & Acc  \\ 
 \hline
 LLaMA-VID$^{\dag}$~\citep{li2023llama}~\pub{ECCV 24}  & {WebVid-Cap} & {VideoChatGPT-100K} & 62.20 & 3.5 & 2.67 & - \\
 Flash-Vstream$^\dag$~\cite{DBLP:journals/corr/abs-2406-08085} & {WebVid-Cap} & {VideoChatGPT-100K} & 65.29 & 3.6 & 2.76 & - \\
 DynFocus ($L=25, K/L=0.8$)  &  WebVid-Cap & VideoChatGPT-100K & 67.90 & 3.7 & 2.91 & 35.1  \\
 \hline
 LLaMA-VID$^\dag$~\citep{li2023llama}~\pub{ECCV 24}  & {WebVid-Cap, LLaVA-CC3M} & {VideoChatGPT-100K, LLaVA-625K} & 68.70 & 3.6 & 2.67 & - \\
LLaMA-VID (Reported)~\citep{li2023llama}~\pub{ECCV 24}  & {WebVid-Cap, LLaVA-CC3M} & {VideoChatGPT-100K, LLaVA-625K} & 69.70 & 3.7 & 2.89 & - \\
 Flash-Vstream$^\dag$~\cite{DBLP:journals/corr/abs-2406-08085} & {WebVid-Cap, LLaVA-CC3M} & {VideoChatGPT-100K, LLaVA-625K} & 69.86 & 3.8 & 2.97 & - \\
 DynFocus ($L=25, K/L=0.8$)   & WebVid-Cap, LLaVA-CC3M & VideoChatGPT-100K, LLaVA-625K & 71.20 & 3.9 & 3.05 & 41.2 \\ 
 \hline
DynFocus ($L=25, K/L=0.8$) & WebVid-Cap, LLaVA-CC3M & + Science-QA & {71.70} & 3.9 & 3.05 & 41.8 \\ 
DynFocus ($L=25, K/L=0.8$) & WebVid-Cap, LLaVA-CC3M & + Science-QA, CLEVRER & {71.60} & 3.9 & 3.07 & 42.6 \\ 
DynFocus ($L=25, K/L=0.8$)  & WebVid-Cap, LLaVA-CC3M & + Science-QA, CLEVRER, NeXT-QA, WebVid-QA & 72.30 & 3.9 & 3.17 & 44.1  \\ 
\hline
\end{tabular}}
\vspace{-.4em}

\label{tab:ablation of data}
\end{table*}

\begin{table*}
\centering
\caption{Performance comparison of existing VideoLLM on VideoHallucer Benchmark for hallucination diagnosis. To evaluate the accuracy, we present the performance of all these models on basic
questions, hallucinated questions, and the overall score. $\dagger$ represents the results by adding rectified prompt $\textit{``Please Carefully Think.''}$, and $\dagger \dagger$ denotes the model with DPO tuning.}
\label{tab:hallu}
\resizebox{1.0\textwidth}{!}{
\begin{tabular}{l|c|cccccccccccccccc} 
\toprule
\multicolumn{1}{c|}{\multirow{2}{*}{\textbf{Models}}}  & \multirow{2}{*}{\begin{tabular}[c]{@{}c@{}}\textbf{LLM}\\\textbf{Size}~~\end{tabular}}  & \multicolumn{3}{c}{\textbf{Object-Relation (\%)}} & \multicolumn{3}{c}{\textbf{Temporal (\%)}} & \multicolumn{3}{c}{\textbf{Semantic Detail (\%)}} & \multicolumn{3}{c}{\textbf{Factual (\%)}} & \multicolumn{3}{c}{\textbf{Non-Factual (\%)}} &  \multicolumn{1}{c}{\multirow{2}{*}{\textbf{Overall}}}\\

\cmidrule(lr){3-5}\cmidrule(lr){6-8}\cmidrule(lr){9-11}\cmidrule(lr){12-14}\cmidrule(lr){15-17}
&   & Basic  & Halluc.  & Final  &  Basic  & Halluc.  & Final & Basic  & Halluc.  & Final  & Basic  & Halluc.  & Final & Basic  & Halluc.  & Final   \\ 
\midrule
VideoChatGPT~\citep{li2023videochat}  & 7B &  \cellcolor{green!20}95.5 &  7.0 & 6.0 & \cellcolor{green!20}100.0 & 0.0 & 0.0 & 96.5 & 4.0 & 2.0 & 86.5 & 13.5 & 7.0 & 85.5 & 27.5 & 17.0 & 6.4 \\
LLaMA-VID~\pub{ECCV 24}~\citep{lin2023video} & 7B & 78.5 & 59.0  & 43.5  & 86.0 & 25.0 & 21.0 & 89.0 & 24.0 & 17.0 & \cellcolor{green!20}98.0 & 2.5 & 2.5 & 16.0 & 14.0 &  3.5 & 21.0 \\
LLaMA-VID~\pub{ECCV 24}~\citep{lin2023video} & 13B & 87.5 & 55.5  & 44.5  & 78.5 & 35.0 & 27.0 & 90.5 & 30.0 & 25.5 & 85.0 & 17.5 & 12.5 & 84.5 & 46.5 & 36.5 & 23.5 \\
Video-LLaMA2~\citep{lin2023video}& 7B & 88.5 & 21.5 & 18.0 & 91.5 & 8.5 & 7.5 &  \cellcolor{green!20}99.0 & 1.5 & 1.0 & 88.0 & 8.5 & 6.5 & 87.5 & 23.5 & 17.0 & 10.0  \\
VideoChat2~\pub{CVPR 24}\citep{li2023mvbench}  & 7B & 26.0 & 41.5 & 10.5 & 23.5 & 25.0 & 7.5 & 33.0 & 26.0 & 9.0 &32.0 & 16.5 & 7.0 & 34.0 & 20.0 & 5.0 & 7.8 \\
VideoLLaVA~\pub{EMNLP 24}~\citep{lin2023video} & 7B & \cellcolor{green!20}95.0 & 38.0 & 34.5 &  \cellcolor{green!20}97.5 & 13.5 & 13.5 &  \cellcolor{green!20}97.0 & 14.0 & 12.0 & 93.0 & 4.5 & 3.0 & 93.0 & 31.5 & 26.0 & 17.8 \\

VideoLaVIT  & - & 94.5 & 39.0 & 35.5 & 88.5 & 27.0 & 25.5 & 96.5 & 13.0 & 10.5 & 97.5 & 6.0 & 4.0 & \cellcolor{green!20}97.5 & 21.5 & 19.0 & 18.9 \\
MiniGPT4-Video~\citep{DBLP:journals/corr/abs-2404-03413} & 7B & 80.5 & 34.5 & 27.5 & 68.5 & 27.0 & 18.0 & 68.5 & 27.0 & 23.5 & 86.0 & 16.5 & 12.0 & 83.5 & 37.5 & 30.5 & 22.3 \\

PLLaVA~\citep{xu2024pllava}  & - & 76.0 & \cellcolor{green!20}76.5 & \cellcolor{green!20}60.0 & 46.5 & \cellcolor{green!20}58.0 & 23.5 & 83.0 & \cellcolor{green!20}71.5 & \cellcolor{green!20}57.0 & 85.0 & \cellcolor{green!20}18.0 & 9.5 & 85.0 & \cellcolor{green!20}53.5 & \cellcolor{green!20}40.5 & \cellcolor{green!20}38.1 \\

LLaVA-NeXT$^{\dagger \dagger}$~\citep{zhang2024llavanextvideo}  & 7B & 72.0 & \cellcolor{green!20}73.0 & 51.5 & 53.0 & \cellcolor{green!20}61.0 & \cellcolor{green!20}28.0 & 63.5 & \cellcolor{green!20}69.0 & 38.0 & 62.5 & \cellcolor{green!20}41.0 & \cellcolor{green!20}14.0 & 61.5 & \cellcolor{green!20}60.5 & 28.5 & 32.0 \\

\hline
DynFocus ($L=25, K/L=0.8$)  & 7B  & 86.5 & 56.0 & 48.0 & 86.0 & 21.5 & 18.5 & 92.0 & 34.0 & 29.0  & 96.5 & 9.0 &  7.5 & - & - & - & - \\
DynFocus$^{\dag}$ ($L=25, K/L=0.8$)  & 7B & 88.0 & 62.0 & \cellcolor{green!20}52.5 & 87.0 & 37.5 & \cellcolor{green!20}{33.5} & 91.5 & 42.0 & \cellcolor{green!20}38.5  & \cellcolor{green!20}98.5 & 15.0 &  \cellcolor{green!20}13.0 & \cellcolor{green!20}96.5 & 40.0 & \cellcolor{green!20}38.5 & \cellcolor{green!20}35.1 \\
\bottomrule
 \end{tabular}
 }
\end{table*}

\subsection{The Effect of Compact Encoding in CCE.} As shown in Table~\ref{tab:CCE}, we introduce several variants to assess the impact of fusion strategies between filtered event prototypes $\mathbf{h}_{t}$ and spatial multi-grained prototypes $\mathbf{G}_{t}$ on model performance. Although direct concatenation uses slightly more tokens compared to cross-attention, it offers performance advantages with greater parameter efficiency, making it our paramount choice. 

\subsection{The Effect of Different Training Datasets}
In this section, we delve into the effect of data scaling on our model. We begin with adopting the only video-based dataset for training. Specifically, we use WebVid-Cap for vision-language alignment in the first stage and VideoChatGPT-100K for instruction tuning in the second stage. Compared with two strong baselines, our model scores 67.9\% on MSVD-QA, even outperforming several models that uses additional image-based dataset for training. As we introduce more image-based dataset, our method consistently shows improving performance, maintaining its leading position. Notably, the addition of CLEVRER appears to degrade the model performance. This possibly because that the visual scene involved in CLEVRER differs significantly from those in the targeted evaluation benchmarks, despite it potentially enhances the spatial reasoning and counting abilities of our model.

\subsection{Different $L$ and $\frac{K}{L}$ towards Long-term Video}
We assess the performance variation with different $L$ and $\frac{K}{L}$ when handling longer and more complex videos, as shown in the following figure,

\begin{figure}[h]
\vspace{-2em}
\centering
 \subfloat[Number of Initial Prototypes]{
\label{fig:l}
\hspace{-1em} 
\includegraphics[scale=0.16]{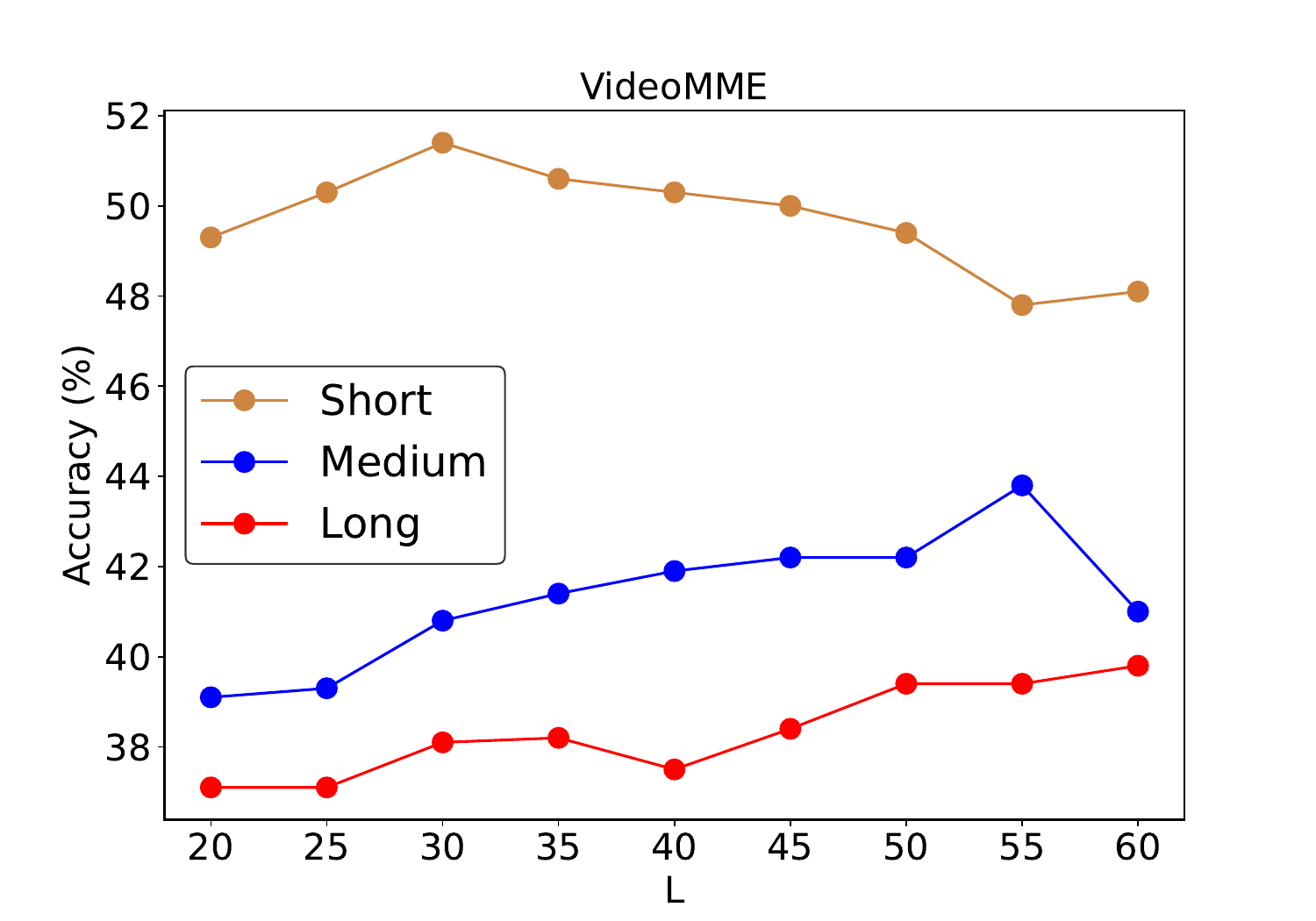}}\hspace{-0.25cm}
\subfloat[Ratio of Filtering Prototypes]{
\label{fig:kl}
\includegraphics[scale=0.16]{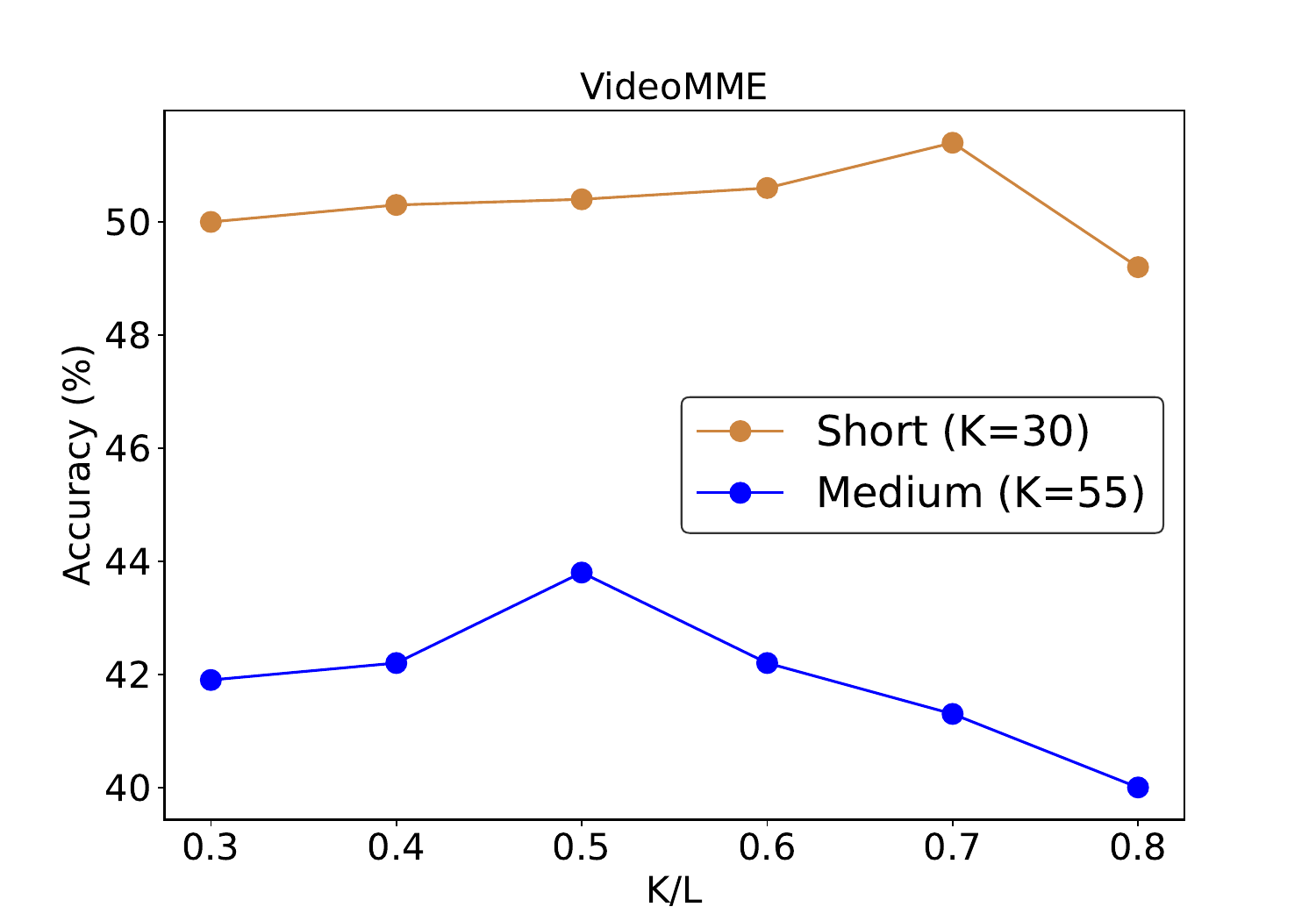}}\vspace{-0.1cm} 
\vspace{-1.3ex}
\end{figure}

\noindent We have two observations for longer videos: (a) the optimal $L$ shifts progressively to the right, from 30 to 55, and further to 60; (b) a smaller $\frac{K}{L}$ yields better performance. This is primarily due to long videos introducing more redundant visual events, while a smaller portion of events should be adaptively selected for question answering. The default parameters towards $L$ and $\frac{K}{L}$ are set to 25 and 0.8 in the main paper when performing evaluation without specification, to achieve a trade-off between accuracy and efficiency. 
\subsection{Robustness on Video Hallucination}
Several researches have pointed that existing MLLMs suffers from the issues of hallucination, which means that they tend to generate irrelevant or nonsensical content that deviates from the original visual context. To comprehensively demonstrate the robustness of our method, we compare the extent of video hallucination of our method with existing video MLLMs. The evaluated benchmark VideoHallucer categorizes hallucinations into two main types: intrinsic and extrinsic, offering further subcategories for detailed analysis, including object-relation, temporal, semantic detail, extrinsic factual, and extrinsic non-factual hallucinations. The overall results are delineated in Table~\ref{tab:hallu}. We have several following observations: (1) Although all models demonstrate strong capabilities in answering basic questions, they experience a significant decline in accuracy when dealing with hallucinated questions. This huge gap implies a widespread conclusion that existing models are vulnerable to the ``Yes/NO' 'bias. In other words, most models tend to generate the ``Yes'' answers. (2) Our \textit{DynFocus} ranks second among all the baselines. VideoChat2 and PLLaVA share the same video-based instructional data but obtain the diametrical results, and the difference stems from source of image-based knowledge. Specifically, the image-based knowledge preserved in PLLaVA originates from a pre-existing image-based MLLM, whereas the knowledge in VideoChat2 is learned from scratch based on collected image QA pairs. On contrary, our model achieves a clear-cut performance gain of \textbf{28.3\%} compared with VideoChat2, and comparable results to PLLaVA. It is noteworthy that our method employs a dynamic encoding strategy, where each frame is encoded with 40 tokens or 2 tokens depending on its contribution to question answering, which is much less than VideoChat2 and PLLaVA.

\section{More Implementation Details}
\label{more_imp}

\subsection{Training Details}
For most of input videos, we sample the frame at 1 \textit{fps} following LLaVA-VID~\cite{li2023llama} and Flash-Vstream~\cite{DBLP:journals/corr/abs-2406-08085}, except excessive long video. All input images or frames are resized to 224 $\times$ 224 and encoded as 16 $\times$ 16 visual features via pre-trained EVA-G~\cite{DBLP:conf/cvpr/FangWXSWW0WC23}, and the hidden dimension $d$ is 1408. We set $I=22$, $J=2$, $P=16$, $K=20$, and $L=25$ when training to achieve a trade-off between performance and memory efficiency. During vision-language alignment, we pre-train our model with a batch size of 256, employing AdamW~\cite{DBLP:journals/corr/KingmaB14} optimizer with a cosine schedule. The learning rate is set to 2e-3, and the warmup rate is 0.03. For instruction tuning, the batch size is 32, and the learning rate is 2e-5. We empirically observe that training more than 1 epoch would hamper performance, we thus set the optimal training epoch to 1. Our model is trained using 8 $\times$ NVIDIA A100 80G GPUs. All training and inference experiments were conducted under BF16 precision to save time and resources. The training settings are summarized in Table~\ref{tab:train_setting}.

\begin{table}[H]
    \caption{Training settings of our \textit{DynFocus.}}
    \label{tab:train_setting}
    \centering
    \begin{tabular}{lcc}
        \toprule
        Settings& Stage-1& Stage-2 \\
        \midrule
        Batch size& 256 & 32 \\ 
        Learning rate& 1e-3 & 2e-5 \\ 
        Learning schedule & \multicolumn{2}{c}{Cosine decay} \\ 
        Warmup ratio & \multicolumn{2}{c}{0.03} \\ 
        Weight decay & \multicolumn{2}{c}{0} \\ 
        Epoch& \multicolumn{2}{c}{1} \\ 
        Optimizer & \multicolumn{2}{c}{AdamW} \\ 
        DeepSpeed stage& 1 & 0 \\ 
        Visual encoder & \multicolumn{2}{c}{Freeze} \\ 
        Projector $\mathcal{F}_{\text{coarse}}, \mathcal{F}_{\text{fine}}$ & \multicolumn{2}{c}{Open} \\ 
        LLM& Freeze & Open \\ 
        
        \bottomrule
    \end{tabular}
\end{table}

\begin{figure*}
    \centering
  \includegraphics[width=1.04\linewidth]{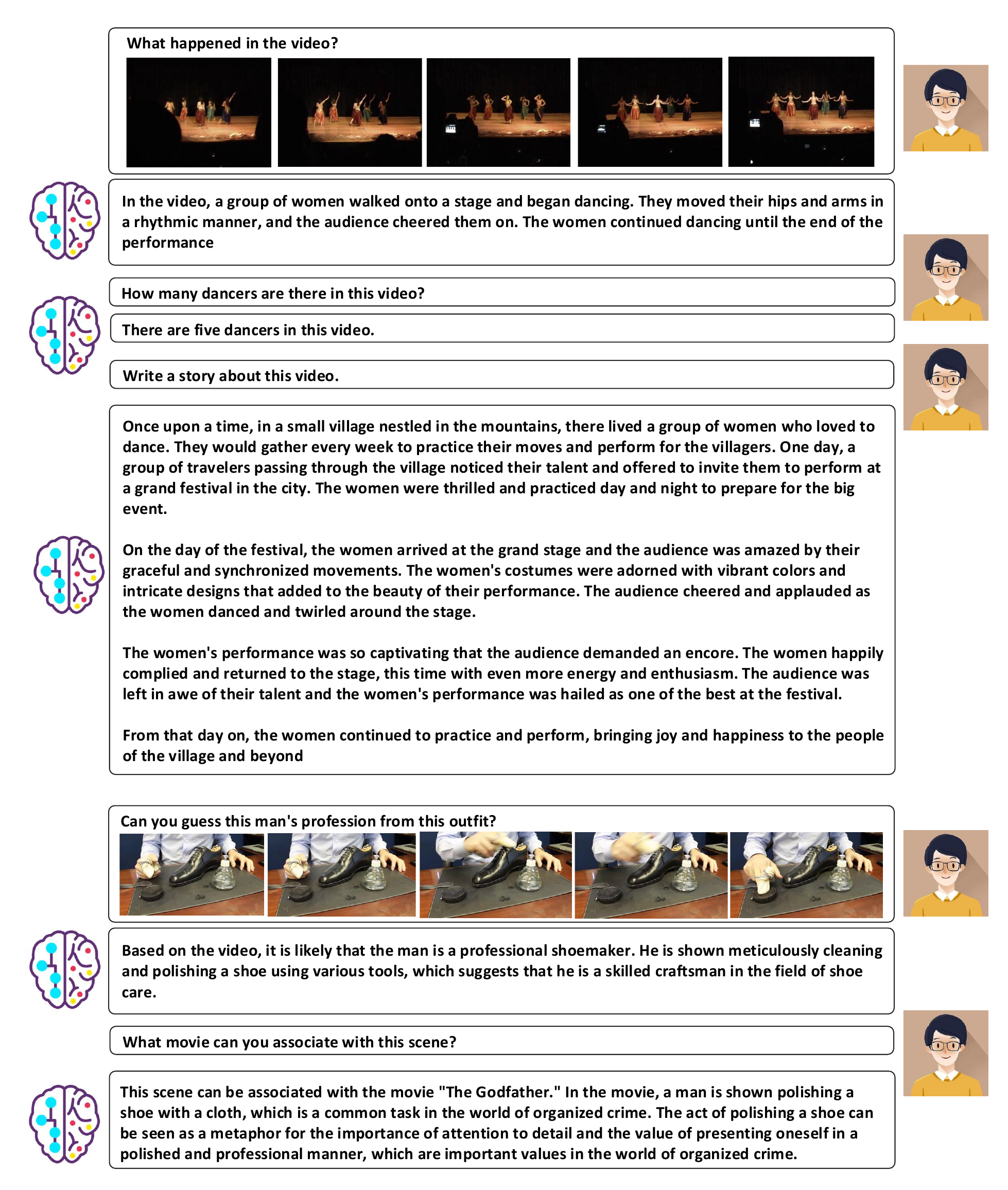}
    \vspace{-2em}
    \caption{Case Studies.}
    \label{fig:case1}
    \vspace{-1em}
\end{figure*}

\begin{figure*}
    \centering
  \includegraphics[width=1.04\linewidth]{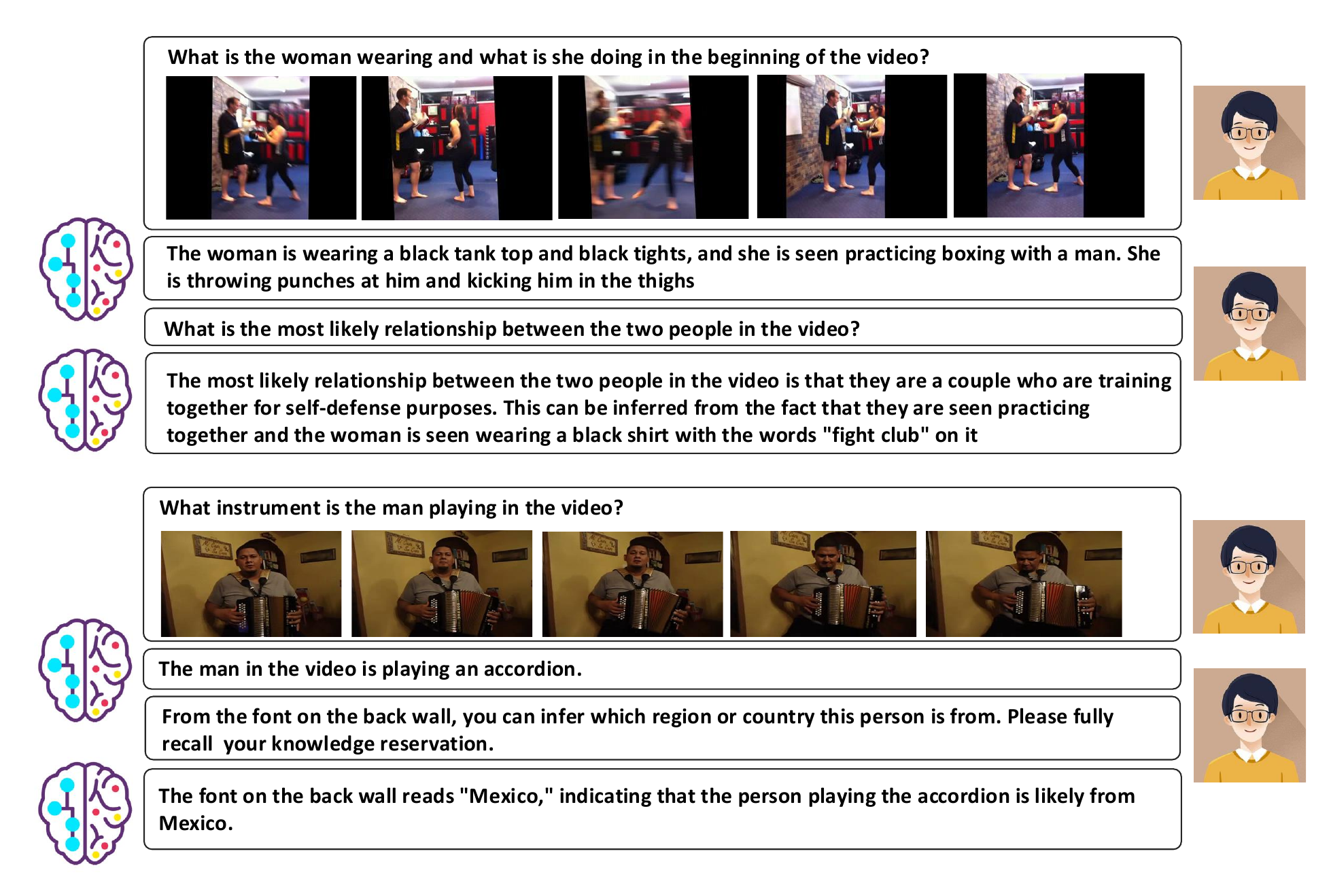}
    \vspace{-2em}
    \caption{Case Studies.}
    \label{fig:case2}
    \vspace{-1em}
\end{figure*}

\subsection{Statistics of Training datasets}
The used training dataset for training are listed in Table~\ref{tab:pretraining} and Table~\ref{tab:tuning}, respectively.

\begin{table}[ht]
\setlength{\tabcolsep}{3pt}
\centering
\caption{Video-Language instructional data statistics for training.}
\vspace{-0.8em}
\scalebox{0.87}{
        \begin{tabular}{c|c|c}
        \hline
        \textbf{Modality} & \multicolumn{1}{c|}{\textbf{Dataset}} & \textbf{Task}   \\ 
        \hline
         & VideoChatGPT~\citep{maaz2023video} & Instruction   \\
         & WebVidQA~\citep{niu2021webvidqa}       & VQA  \\
         & CLEVRER~\citep{yi2019clevrer} & VQA   \\
        \multirow{-4}{*}{Video-Text} & NeXT-QA~\citep{bain2021frozen} & VQA   \\ 
        \hline
         & COCO~\citep{lin2014microsoft} & Captioning   \\
         & Visual Genome~\citep{krishna2017visual} & Captioning   \\
         & GQA~\citep{hudson2019gqa} & VQA   \\ 
         & OCRVQA~\citep{verma2018ocrvqa} & VQA   \\
         & TextVQA~\citep{agrawal2019textvqa} & VQA  \\

        \multirow{-6}{*}{Image-Text} & ScienceQA~\citep{li2021scienceqa} & VQA                    \\ 
        \hline
        Vision-Language & Total & Mixture   \\
        \hline
        \end{tabular}}
\label{tab:pretraining}
\end{table}

\begin{table}[ht]
\setlength{\tabcolsep}{3pt}
\centering
\caption{Video-Language pre-training data statistics for training. We directly adopt the filtered version following LLaVA-VID~\cite{li2023llama}.}
\vspace{-0.8em}
\scalebox{0.87}{
        \begin{tabular}{c|c|c}
        \hline
        \textbf{Modality} & \multicolumn{1}{c|}{\textbf{Dataset Source}} & \textbf{Task}   \\ 
        \hline
        \multirow{-1}{*}{Video-Text} & WebVid-Cap~\citep{DBLP:conf/iccv/BainNVZ21} & Captioning  \\ 
        \hline
        \multirow{-1}{*}{Image-Text} & LLaVA-filtered CC3M~\citep{DBLP:conf/acl/SoricutDSG18} & Captioning                  \\ 
        \hline
        Vision-Language & Total & Captioning   \\
        \hline
        \end{tabular}}
\label{tab:tuning}
\vspace{-1em}
\end{table}

\subsection{Details of Long-Term Video Benchmark}

\textbf{LV-Bench.} It encompasses a diverse set of tasks aimed at long video comprehension and information extraction, which tests six core capabilities. Temporal Grounding (TG) focuses on understanding sequences and dynamics within the video. Summarization (Sum) requires an entire understanding of video from start to finish. Reasoning (Rea) involves four advanced reasoning skills: casual relationship identification, understanding for emotional development of character, understanding for underlying intentions of characters, future prediction. Entity Recognition (ER) requires the key entities tracking (such as people, places, and objects) throughout the video. Event Understanding (EU) needs to summarize the semantic concept for question answering. Key Information Retrieval (KIR) emphasizes retrieval of crucial detailed clues within videos.

\noindent$\textbf{MLVU}.$ The evaluation task of MLVU can be categorized into three types: (1) \textit{holistic LVU} (TR: Topic Reasoning, AR: Anomaly Recognition, VS: Video Summary), which requires to make use of global perspectives from the entire video; (2) \textit{single-detail LVU} (NQA: Needle QA, ER: Ego Reasoning, PQA: Plot QA, SSC: Sub-Scene Captioning), which needs to pinpoint one critical details in a haystack; (3) \textit{multi-detail LVU} (AO: Action Order, AC: Action Count), which calls for the joint utilization of multiple detailed plots within the long video to collaborately infer the answer. 

\begin{figure}[ht]
    \centering
    \includegraphics[width=\linewidth]{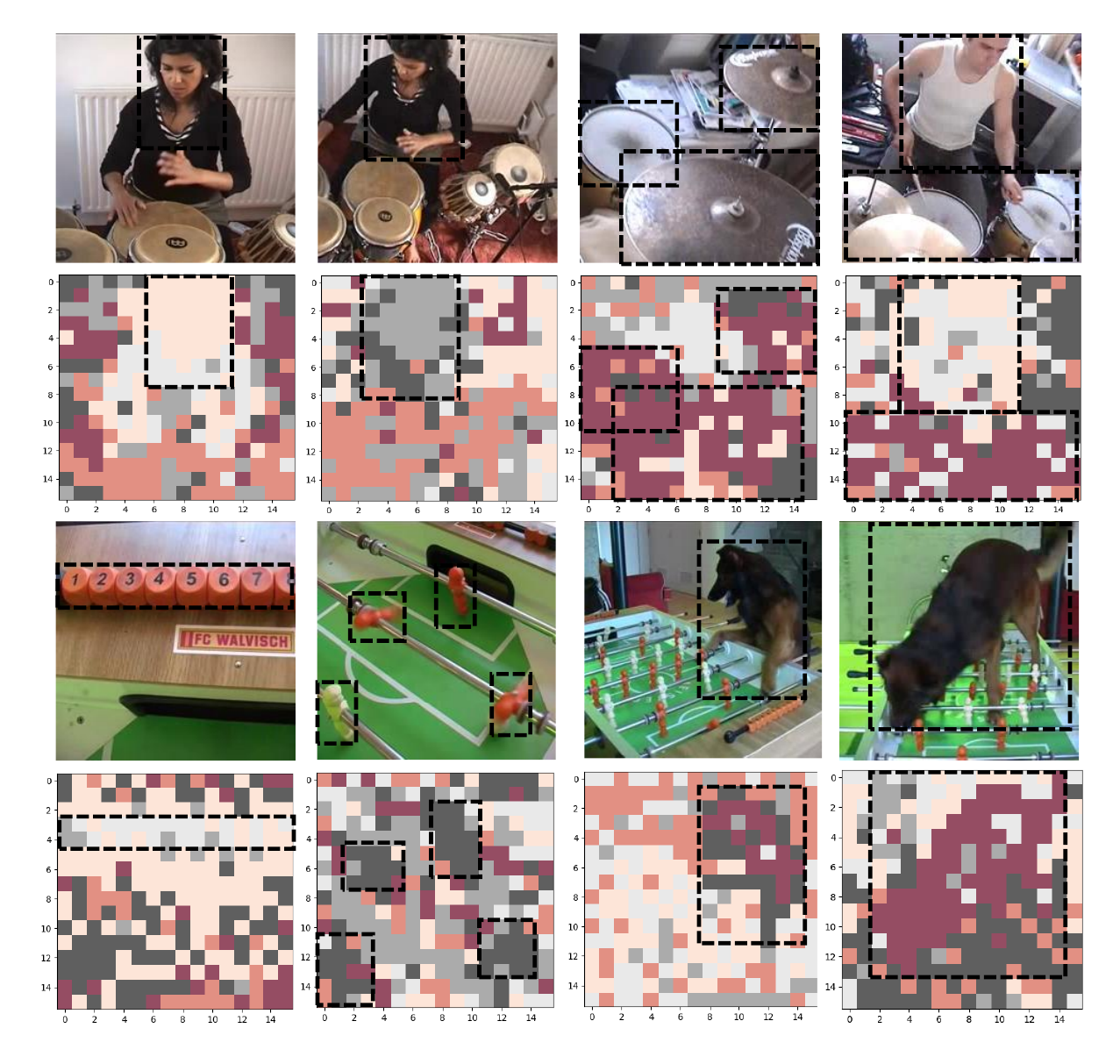}
    \vspace{-2em}
    \caption{Illustration of learned spatial prototypes in S-DPC. We highlight the region with dotted line for better correspondence.}
    \label{fig:dpcknn}
    \vspace{-1em}
\end{figure}

\section{More Visualization Results}
\label{more_vis}
In Figure~\ref{fig:dpcknn}, we illustrate the learned semantic prototypes, where the patches with similar semantic are first clustered. The formation of spatial prototypes effectively reduces the token number while enhancing the semantic representation of each video frame.


\section{Case Study}
Figure~\ref{fig:case1} and Figure~\ref{fig:case2} illustrates the conversation example towards video understanding. Our method could harness the information of contextual clues to  provide appropriate and coherent responses based on user prompts. The illustrative examples showcase the remarkable ability of $\textit{DynFocus}$ on capturing the temporal dynamics and delicate visual details, addressing the counting problem as well as imagination across multiple conversational turns.
\label{case}

\end{document}